%% file: main.tex
\pdfoutput=1
\documentclass[11pt]{article}

\usepackage[preprint]{acl}
\usepackage{array}

\usepackage{times}
\usepackage{latexsym}

\usepackage[T1]{fontenc}
\usepackage[utf8]{inputenc}

\usepackage{microtype}

\usepackage{inconsolata}
\usepackage{graphicx}
\usepackage{algorithm}
\usepackage{algpseudocode}
\usepackage{multirow}
\usepackage{xcolor}
\usepackage{color, colortbl}
\definecolor{LightCyan}{rgb}{0.88,1,1}
\usepackage{tabularx}
\usepackage{paralist}
\usepackage{subcaption}
\usepackage{longtable}
\usepackage{amssymb}% http://ctan.org/pkg/amssymb
\usepackage{pifont}% http://ctan.org/pkg/pifont
\usepackage{makecell}
\newcommand{\cmark}{\ding{51}}%
\newcommand{\xmark}{\ding{55}}%
\newcommand{\thickhline}{\noalign{\hrule height 1.5pt}}
\usepackage{hyperref}

\hypersetup{
    colorlinks=true,
    linkcolor=red,     % Color of internal links (sections, pages, etc.)
    citecolor=green,    % Color of citation links
    filecolor=magenta,  % Color of file links
    urlcolor=blue       % Color of external links
}

\title{Solving for X and Beyond: Can Large Language Models Solve Complex Math Problems with More-Than-Two Unknowns?}%Do Large Language Models' Capabilities Extend to Complex Math Problems with More-Than-Two Unknowns?}

\author{Kuei-Chun Kao$^1$, Ruochen Wang$^1$, Cho-Jui Hsieh$^1$ \\
  $^1$Department of Computer Science, University of California, Los Angeles \\
  \texttt{\{johnson0213, ruocwang\}@g.ucla.edu} \\
  \texttt{chohsieh@cs.ucla.edu} \\ \\
  \href{https://johnsonkao0213.github.io/Formulate\_and\_Solve/}{https://johnsonkao0213.github.io/Formulate\_and\_Solve/}}

\begin{document}
\maketitle
\begin{abstract}
% Large Language Models (LLMs) have demonstrated strong performance on many mathematical reasoning datasets, but these datasets are often limited to problems with one or two unknowns.
% To address this, we introduce BeyondX, the first multi-unknown algebraic benchmark comprising 480 problems with three, four, and five unknowns.
% We developed a novel LLM-based pipeline to automatically expand existing problems, overcoming the difficulty of manually creating such datasets.
% A comprehensive evaluation of current LLMs, including general and fine-tuned models with existing prompting techniques, reveals a significant performance drop when facing an increased number of unknowns.
% For instance, GPT-4 with Zero-shot-CoT achieves near 90\% accuracy on problems with up to two unknowns but only 20\% on problems with five unknowns.
% To investigate whether this performance gap is due to insufficient prompting techniques, we introduce Formulate-and-Solve, an automated prompting method that enables solving problems with an arbitrary number of unknowns.
% Our empirical evaluations show that Formulate-and-Solve outperforms traditional prompting methods on both standard algebra problem datasets and the more challenging BeyondX dataset.
% Formulate-and-Solve not only improves LLM performance but also provides clearer insights into the actual limitations of current models when faced with advanced mathematical challenges.
Large Language Models (LLMs) have demonstrated remarkable performance in solving math problems, a hallmark of human intelligence.
Despite high success rates on current benchmarks; however, these often feature simple problems with only one or two unknowns, which do not sufficiently challenge their reasoning capacities.
This paper introduces a novel benchmark, BeyondX, designed to address these limitations by incorporating problems with multiple unknowns.
Recognizing the challenges in proposing multi-unknown problems from scratch, we developed BeyondX using an innovative automated pipeline that progressively increases complexity by expanding the number of unknowns in simpler problems.
Empirical study on BeyondX reveals that the performance of existing LLMs, even those fine-tuned specifically on math tasks, significantly decreases as the number of unknowns increases - with a performance drop of up to 70\% observed in GPT-4.
To tackle these challenges, we propose the Formulate-and-Solve strategy, a generalized prompting approach that effectively handles problems with an arbitrary number of unknowns.
Our findings reveal that this strategy not only enhances LLM performance on the BeyondX benchmark but also provides deeper insights into the computational limits of LLMs when faced with more complex mathematical challenges.
\end{abstract}

\newcommand{\wrc}[1]{\textcolor{blue}{Ruochen \scriptsize: #1}}

\input{sections/01-Introdution}

\input{sections/02-Related-Work}
\input{sections/03-Math-Word-Problem-Generation}

\input{sections/04-Preliminary}
\input{sections/05-Math-Word-Problem-Solver}

\input{sections/06-Experiment}

\input{sections/07-Discussion}
\input{sections/08-Conclusion}

\input{sections/09-Limitations}

% \section*{Acknowledgements}

% Additional elements were taken from the formatting instructions of the \emph{International Joint Conference on Artificial Intelligence} and the \emph{Conference on Computer Vision and Pattern Recognition}.

\bibliography{custom}

\input{sections/10-Appendix}

\end{document}

%% file: sections/01-Introdution.tex
\section{Introduction}
Mathematical problem-solving is a fundamental aspect of human intelligence, necessitating both language comprehension and reasoning skills.
Recently, LLMs pretrained on extensive web-scale datasets, have exhibited exceptional abilities in addressing a variety of complex tasks.
Consequently, mathematical challenges are frequently employed to benchmark the reasoning abilities of LLMs.
Studies have shown that these models demonstrate human-level efficacy in solving math problems, aided by diverse prompting techniques include in-context learning \cite{wei2022chain, kojima2022large, wang2022self} and the integration of external computational tools~\cite{gao2022pal, chen2022program, liu2023plan, he2023solving}.

Existing math datasets (see Table~\ref{tab:dataset}) used to evaluate LLMs often consist of algebraic problems involving only one or two unknown variables.
While current results on these datasets are promising, their simplicity masks the true capabilities and limitations of these models.
For instance, GPT-4~\cite{achiam2023gpt} achieves a 98\% success rate on the GMS8K~\cite{cobbe2021training} dataset, suggesting that performance on these benchmarks is nearing saturation.
This highlights the need for the development of more complex problem sets designed to rigorously stress test LLMs and provide a more accurate measure of their performance.

While quantifying the complexity of these math problems is multi-dimensional, one common measure is the number of unknowns required to solve the problem.
Problems with more unknowns involve larger systems of equations, reflecting more complex relationships between the quantities, and thus demanding more sophisticated solving methods.
However, creating datasets that include problems with multiple unknowns presents significant challenges, as it is difficult for humans to manually develop a sufficient number of these complex problems from scratch.
As a result, \textbf{existing math datasets are dominated by problems with at most two unknowns}~\cite{cobbe2021training, koncel-kedziorski-etal-2016-mawps, koncel2015parsing, roy2018mapping}.

This paper tackles the aforementioned challenge systematically, by presenting three key contributions:
(1) the development of a multi-unknown math benchmark,
(2) an empirical study assessing the performance of current LLMs on this new benchmark, and
(3) the introduction of a specialized prompting strategy designed to enhance the ability of LLMs to solve multi-unknown problems.

\paragraph{C1: BeyondX - The first multi-unknown algebraic benchmark.}
To efficiently generate a large corpus of multi-unknown problems, we developed a novel pipeline that automatically expands existing problems to $N$ unknowns.
This pipeline operates on three key ideas:
(1). \textit{Scenario Expansion}: New problems are derived by extending the scenarios (such as financial calculations or grocery shopping) of existing simpler problems, ensuring contextual relevance.
(2). \textit{Progressive Extrapolation}: We add unknowns incrementally — one at a time — progressing from problems with two unknowns to three, four, and so on.
This step-by-step approach makes the problem generation process significantly more manageable.
(3). \textit{Decomposed Problem Generation}: Instead of creating an entire problem at once, we decompose the process.
The LLM is carefully instructed to first introduce new unknown variables into the base scenarios, generate the corresponding equations, translate these equations into semantic statements, and finally integrate them into the comprehensive problem statement.

\paragraph{C2: Existing LLMs struggles with increasing unknowns.}
Utilizing our BeyondX benchmark, we conduct a comprehensive evaluation of current LLMs, which includes both general-purpose models like GPT-3.5~\cite{brown2020language}, GPT-4, Gemini-Pro~\cite{team2023gemini}, and Mistral-7B~\cite{jiang2023mistral} as well as models specifically fine-tuned on mathematical problems (WizardMath~\cite{luo2023wizardmath}, OpenMath~\cite{toshniwal2024openmathinstruct}, and MetaMath~\cite{yu2023metamath}).
Our findings reveal a significant drop in performance as the number of unknowns in problems increases - a staggering $\sim 70\%$ degradation on GPT-4 for instance.

\begin{compactitem}
    \item This marked decline indicates that current benchmarks may overstate the algebraic capabilities of these models.
    \item Additionally, despite efforts to fine-tune LLMs on previous math corpus, they still struggle with multi-unknown problems.
    \item Even sophisticated prompting strategies, which utilize detailed natural language explanations, fail to substantially aid LLMs in overcoming these more complex challenges.
\end{compactitem}

\paragraph{C3: Formulate-and-Solve A prompting method to tackle multi-unknown problems.}
Traditional prompting methods for LLMs typically do not account for the complexity inherent in systems of equations, potentially limiting the math-solving capabilities of these models.
Addressing whether the observed performance drop in LLMs is primarily due to inadequate prompting strategies forms a critical part of future investigation.
As the initial step, we propose Formulate-and-Solve, an automated prompting method that generalizes to an arbitrary number of unknowns.
This strategy refines current approaches by integrating general math-solving principles to automatically craft relevant multi-unknown in-context examples for LLMs.

Our empirical evaluations demonstrate that Formulate-and-Solve outperforms traditional prompting methods on both standard algebra problem datasets and our more challenging BeyondX dataset.
Importantly, our findings suggest that while the inherent limitations of LLMs contribute to their underperformance on complex problems, inadequate prompting strategies are a substantial bottleneck.
By enhancing these strategies, Formulate-and-Solve not only improves LLM performance but also provides clearer insights into the actual computational limitations of current models when faced with advanced mathematical challenges.

%%%%%%%%%%%%%%%%%%%%%%%%%%%%%%%%%%%%%

% \begin{figure}[t]
% \centering
% \includegraphics[width=\columnwidth]{latex/figures/mwp_3.png}
%    \caption{An example of multi-unknown algebra problem and its underlying system of equations and variable meaning.}
% \label{fig:mwp}
% \end{figure}

% \input{latex/tables/dataset_overview}

%% file: sections/02-Related-Work.tex
\section{Related Work}

\subsection{Math Word Problem Generation}
Early research on math word problem (MWP) generation relied heavily on pre-defined structures, including domain knowledge, equations, and text templates~\cite{nandhini2011math, williams2011generating, polozov2015personalized}.
More recently, researchers began using pre-trained models fine-tuned on equation-to-MWP examples~\cite{wang-etal-2021-math}. Studies on using LLMs for MWP generation are scarce. 
Existing work includes evaluating GPT-3's ability to mimic specific problem types~\cite{zong2023solving} and using GPT-4 to improve readability in existing problems~\cite{norberg2023rewriting}.
However, these approaches are limited to replicating existing problem structures, such as the number of unknowns or equation templates. Our work focuses on how to expand existing one or two unknown problems into more complex multiple unknowns.

\subsection{Math Word Problem Solver}
Mathematical reasoning skills are crucial for intelligent systems, leading to a surge in research. In the past, studies focused on how statistical and deep learning NLP models could solve arithmetic and algebraic problems~\cite{hosseini2014learning, koncel2015parsing, roy2016solving, liang2016tag, ling2017program}. Recently, researchers have introduced increasingly challenging math datasets~\cite{saxton2019analysing, amini2019mathqa, miao-etal-2020-diverse, patel2021nlp, hendrycks2021measuring} to improve difficulty, diversity, and robustness. However, these methods struggle to adapt to problems outside their training data. This limitation has driven the rise of LLMs in mathematical reasoning.

\subsection{Math Reasoning with LLMs}
%As the field of LLMs continues to prosper, 
Many prompting techniques have emerged to unlock the reasoning abilities of LLMs~\cite{qiao2022reasoning}. 
Chain-of-Thought (CoT) Prompting \cite{wei2022chain, kojima2022large, wang2022self} was proposed to generate the reasoning steps before submitting the answer. Later, several other works ~\cite{nye2021show, zhou2022least, drozdov2022compositional, wang-etal-2023-plan} also proposed different approaches to utilize LLMs to solve reasoning tasks by allowing intermediate steps or planning first before solving. All of these methods allow LLMs to process all steps without using any external tools or refinements. For incorporating external tools, Programming-of-Thought (PoT) prompting~\cite{chen2022program, gao2022pal} utilizes LLMs with code pretraining to write a program as a rationale that explains the reasoning process. Equation-of-Thought (EoT)~\cite{liu2023plan, he2023solving} prompting tackles MWPs by converting them into linear equation systems, which are then solved by a symbolic solver. Although PoT and EoT tried to use external tools to solve MWPs, they did not consider the scenario of multiple unknown variables.

Recent LLMs advancements for math reasoning involve various training approaches. One method focuses on pretraining data specifically designed for math, such as Minerva~\cite{lewkowycz2022solving}, Llemma~\cite{azerbayev2023llemma}, and DeepSeekMath~\cite{shao2024deepseekmath}. Another approach involves generating synthetic questions and answers that mimic existing benchmarks. For example, the WizardMath series~\cite{luo2023wizardmath} that improves mathematical reasoning in Mistral 7B~\cite{jiang2023mistral} with problems sourced primarily from GSM8K and MATH~\cite{hendrycks2021measuring} via output from closed-source LLMs. 
MetaMath~\cite{yu2023metamath} and MMIQC~\cite{liu2024augmenting} focus on expanding existing questions in GSM8K and MATH. MetaMath rewrites questions in various ways, while MMIQC combines existing math pretraining data such as OpenWebMath~\cite{paster2023openwebmath} with question-answer variations from MetaMath. 
The Mammoth series (including Mammoth2)~\cite{yue2023mammoth, yue2024mammoth2} uses curated instruction tuning datasets (MathInstruct, WebInstrcut) with reasoning rationales for training. The OpenMathInstruct~\cite{toshniwal2024openmathinstruct} series utilizes synthetic instruction data from open-source LLMs with strong math reasoning abilities.

%related work minimize

%% file: sections/03-Math-Word-Problem-Generation.tex
\section{Automatic Generation of Multi-Unknown Algebra Problems via Progressive Expansion} \label{sec:generation}
% This section layouts our methods for creating BeyondX Benchmark.
% Section~\ref{} describes the challenges of 
% Section~\ref{} discusses
% Section~\ref{} introduces

\begin{figure*}[t]
\centering
\includegraphics[width=0.8\textwidth]{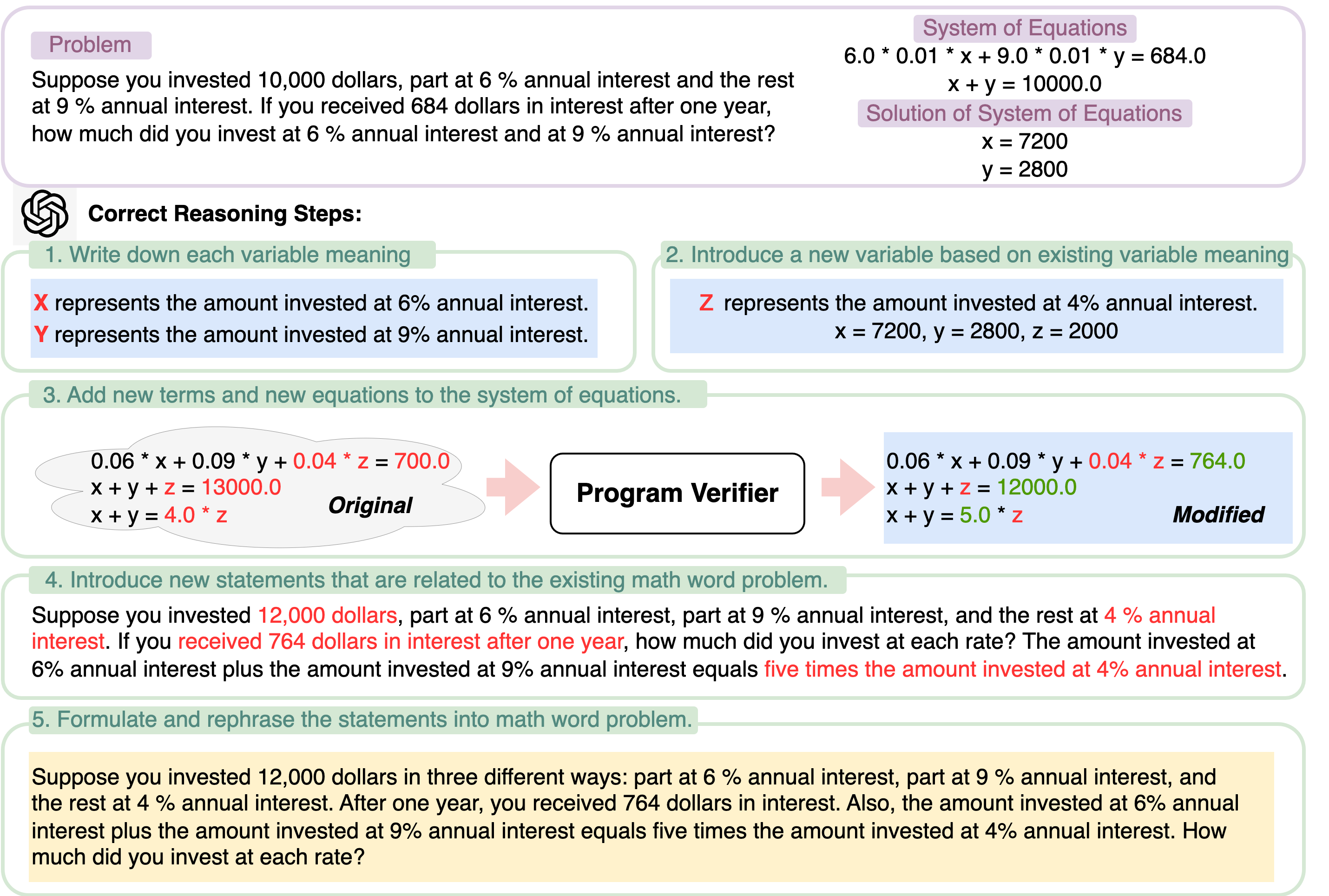}
\vspace{-2mm}
\captionsetup{font=small, width=0.9\linewidth}
\caption{An example question of multi-unknown algebra problem generation and its corresponding reasoning steps. The prompts used for each step can be found Appendix~\ref{tab:mwp_generation_instruction}.}
\label{fig:mwp_generation_fig}
\vspace{-4mm}
\end{figure*}

\subsection{Challenges for Constructing Multi-Unknown Datasets}
\paragraph{Generating new problems with LLMs.}
Creating correct, diverse, and solvable math problems manually is an exceptionally laborious task.
The complexity of this task increases with the addition of each unknown, as more unknowns require consideration of additional relationships within the problem scenario.
To automatize this process, we employ LLMs to generate the problems, with human verifiers subsequently ensuring the quality and solvability of these problems.

\paragraph{Limitations of naive generation.}
Directly prompting LLMs to generate multi-unknown algebra problems has often resulted in poor quality outputs.
Firstly, generating problem scenarios from scratch tends to produce a narrow range of problem types, as evidenced by the lack of diversity reported in Table~\ref{tab:direct_mwp_generation}.
Secondly, attempting to generate all relevant relationships and corresponding equations in a single step frequently leads to violations of problem constraints, rendering many problems unsolvable as detailed in Table~\ref{tab:equation_aware_mwp_generation_part1}.

\subsection{Generating New Problems via Progress Expansion} \label{sec:generation_steps}
\paragraph{Pipeline overview.}
To address the aforementioned challenges, we introduce a novel approach called Progressive Expansion, which applies a divide-and-conquer strategy:
(1). \textit{Scenario Diversification:} We begin by expanding existing simpler problems to increase scenario diversity. This leverages the rich variety of simpler problem scenarios as a foundation for more complex questions.
(2). \textit{Incremental Expansion:} Instead of expanding problems from 1-2 unknowns to $N$ unknowns in a single leap, we incrementally introduce one new variable at a time. This step-by-step approach simplifies the transformation from $N-1$ unknowns to $N$, making it more manageable and controllable for LLMs.
(3). \textit{Enhanced Solvability:} The problem expansion is broken into several simpler stages, making the entire generation more tractable for LLMs.

\paragraph{Multi-step problem expansion.}
The process of expanding problems is systematically divided into five steps, as illustrated in Figure~\ref{fig:mwp_generation_fig}.
\textit{Step 1: Understanding the source problem.}
First, we instruct the LLM to analyze the original problem (including solutions) and explicitly explain the role of each unknown variable. This ``perception step" lays the groundwork for subsequent expansions.
\textit{Step 2: Introducing a new unknown.}
The LLM then introduces an additional unknown variable related to the existing problem framework and assigns an appropriate initial value (oracle value) to this variable.
\textit{Step 3: Expanding equation sets.}
Next, the LLM generates a new equation that delineates the quantitative relationships between the new and existing variables. To ensure these equations are solvable, we integrate a Program Verifier module to assess and adjust their correctness as needed.
\textit{Step 4: Add equations to the problem statement}
We translate the newly formed equations into text and incorporate them into the original problem statement to maintain consistency and flow.
\textit{Step 5: Final refinement.}
Finally, we engage the LLM in a thorough polishing phase to refine the problem statement, ensuring it is fluent and coherent.

\subsection{Constructing the Benchmark}

\paragraph{Seed problems.}
We select ALG514~\cite{kushman-etal-2014-learning} and DRAW-1K~\cite{upadhyay-chang-2017-annotating} as the foundational seed problems to expand.
These datasets are particularly suitable as they include full solutions with oracle equation sets.

\paragraph{Statistics.}
With this generation process, we selected a total of 464 problems. Specifically, there are 194 problems with three unknowns, 158 problems with four unknowns, and 112 problems with five unknowns. In addition, since our generated dataset is expanded from the existing dataset, it contains various topics or subjects including moving objects, liquids, interest, distance, and geometry.

%% file: sections/04-Preliminary.tex
\section{Benchmarking existing LLMs and Prompting Methods}

\begin{figure*}[tp]
 \centering
  \begin{subfigure}{0.32\textwidth}
    \includegraphics[width=\linewidth]{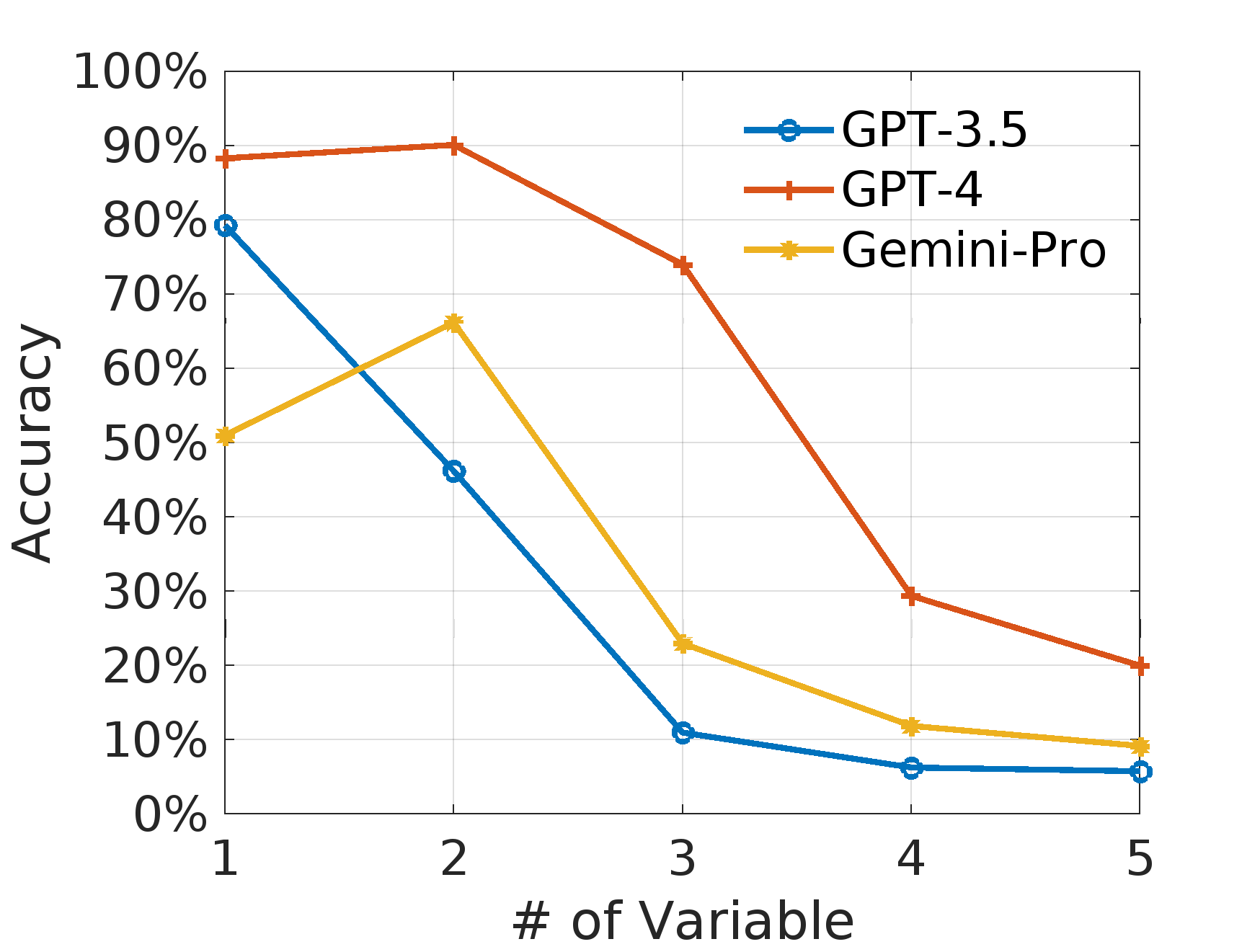}
    \captionsetup{font=small, width=0.8\textwidth} % Adjust the width of the caption
    \caption{Zero-shot-CoT performance of Close-source model.}
    \label{fig:closed_source_model}
  \end{subfigure}
  \begin{subfigure}{0.32\textwidth}
    \includegraphics[width=\linewidth]{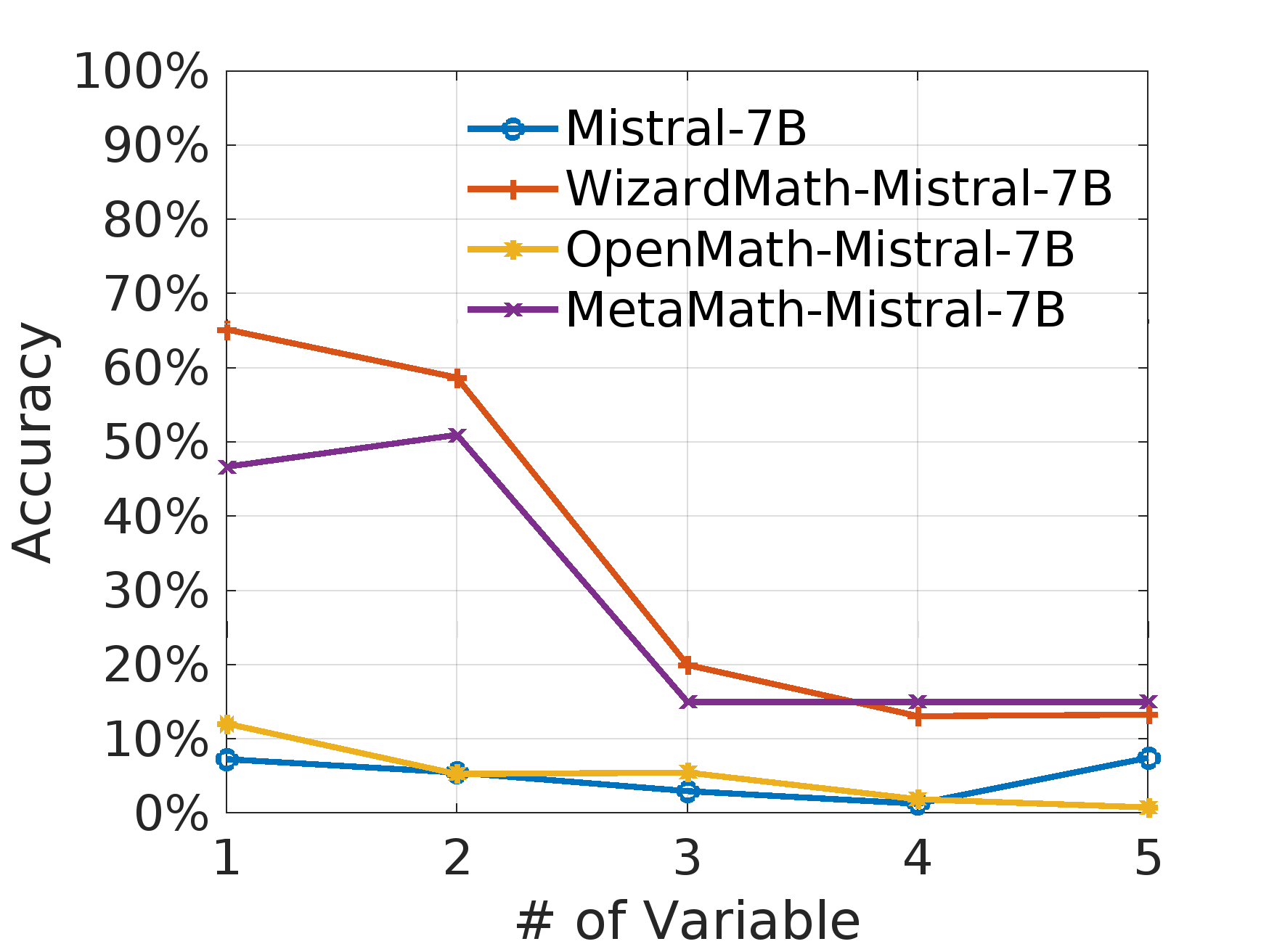}
    \captionsetup{font=small, width=0.8\textwidth} % Adjust the width of the caption
    \caption{Zero-shot-CoT performance of Open-source model.}
    \label{fig:open_source_model}
  \end{subfigure}
  \begin{subfigure}{0.32\textwidth}
    \includegraphics[width=\linewidth]{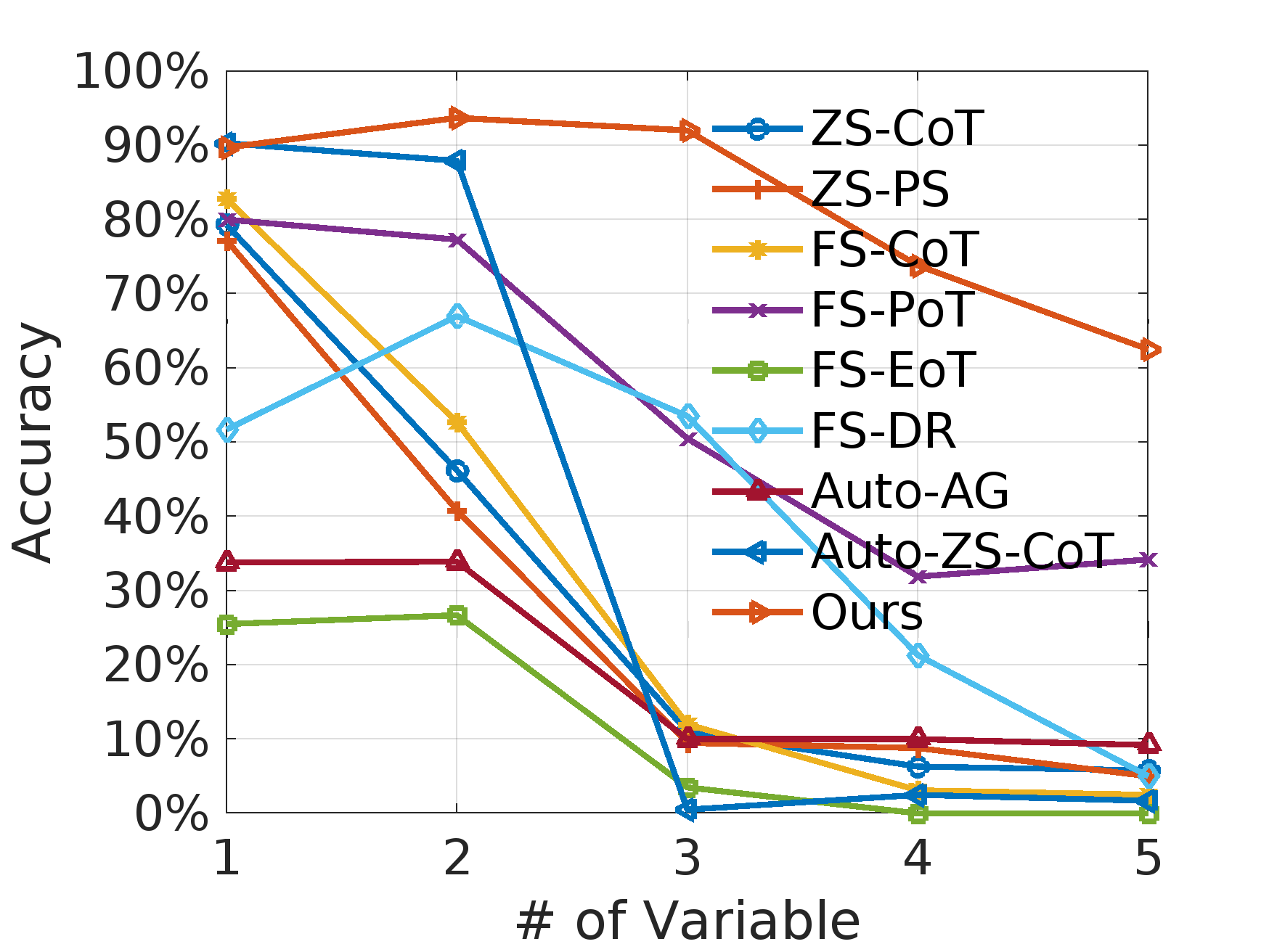}
    \captionsetup{font=small, width=0.8\textwidth} % Adjust the width of the caption
    \caption{Different prompting methods performance of GPT-3.5.}
    \label{fig:different_method_alg514}
  \end{subfigure}
\vspace{-2mm}
\captionsetup{font=small, width=0.9\linewidth}
\caption{Preliminary study of different LLMs and prompting methods on multi-unknown algebra datasets.}
\vspace{-5mm}
\label{fig:LLM_zero_shot_perf}
\end{figure*}
% ZS- means zero-shot, FS- means few-shot with manual demonstrations, and Auto- means few-shot with automatic demonstrations.

% We conduct a comprehensive study evaluating the performance of existing LLMs and prompting methods on multi-unknown problems. 

\subsection{LLMs for solving multi-unknown algebra problems}
To evaluate the performance of various LLMs on BeyondX, we consider the Zero-shot-CoT prompting method (details in Section~\ref{Sec:exp_setting}) and test the performance of both General-Purpose LLMs (GPT-3.5, GPT-4, Gemini-Pro, Mistral-7B) and Mathematically fine-tuned LLMs (WizardMath, OpenMath, MetaMath).
%test the following models with Zero-shot-CoT prompting: \\
%\textbf{\emph{General-Purpose LLMs.}} GPT-3.5, GPT-4, Gemini-Pro, and Mistral. \\
%\textbf{\emph{Mathematical LLMs.}} WizardMath, OpenMath, and DeepSeekMath.

In Figure~\ref{fig:closed_source_model} and Figure~\ref{fig:open_source_model}, the results show a significant performance drop with multiple unknowns on both closed-source and open-source LLMs. For example, GPT-4 achieves near 90\% accuracy when solving problems with 1 or 2 unknowns, but the performance drops to 20\% when solving problems with 5 unknowns. This highlights that current LLMs with Zero-shot-CoT are not able to solve multi-unknown problems, and this limitation was not recognized in the literature due to lack of datasets. 
In addition, mathematically fine-tuned LLMs exhibit a significant performance drop when encountering problems with more than two unknowns.
This reveals a limitation of current fine-tuning methods, highlighting the need for improved algorithms or training sets.

\subsection{Prompting Methods}
Figure~\ref{fig:closed_source_model} demonstrates that state-of-the-art LLMs cannot solve multi-unknown problems with Zero-shot-CoT prompting. To investigate whether this issue can be mitigated with better prompting methods, we evaluated nine existing prompting methods using GPT-3.5, categorized into three types:  \\
\textbf{\emph{Zero-shot.}} Zero-shot-CoT~\cite{kojima2022large} and Plan-and-Solve~\cite{wang-etal-2023-plan} prompting. \\
\textbf{\emph{Few-shot with manual demonstrations.}} CoT~\cite{wei2022chain}, PoT~\cite{gao2022pal}, EoT~\cite{liu2023plan}, and Declarative~\cite{he2023solving} prompting. \\
\textbf{\emph{Few-shot with automatic demonstrations.}} Analogical~\cite{yasunaga2023large}, Auto-Zero-shot-CoT.

In Figure~\ref{fig:different_method_alg514}, we observe zero-shot and few-shot CoT prompting methods seem inadequate when solving multi-unknown problems. We find that while CoT correctly sets up the equations, it fails to accurately solve the system of equations.
Additionally, even though some prompting methods like PoT, EoT, and Declarative use external tools as a calculator and equation solver, they manually design their demonstration for simpler problems and fail to generalize to more complex multiple unknown scenarios.
Although some methods construct demonstrations automatically from the problem context (Analogical, Auto-Zero-shot-CoT), they still suffer from poor performance.
%We also observe that even if some methods guide LLMs to automatically generate demonstrations from the problem context, they still cannot perform well on multi-unknown problems. 
Since LLMs themselves do not have enough capability to solve multi-unknown problems, the generated demonstrations are often of low quality. This raises concerns about prompt engineering requiring "human-in-the-loop" solutions with domain knowledge integrated through instructions. 

Therefore, in the next section, we will go through a detailed formulation of Formulate-and-Solve prompting. Our method can significantly bridge the gap as shown in Figure \ref{fig:different_method_alg514}.
% In comparison, our proposed method \textbf{Modeling-and-Solve} manages to solve the problem automatically using a system of equations as an intermediate form.

%% file: sections/05-Math-Word-Problem-Solver.tex
\section{Automatic Solver of Algebra Problems} \label{sec:solver}

%Our method of addressing these demanding questions in the BeyondX is comprised of several distinct modules, as outlined in Figure~\ref{fig:mwp_solver_fig}.
To investigate whether the observed performance drop is primarily due to inadequate prompting strategies or the limitation of LLMs. we develop \textbf{Formulate-and-Solve}, an automated prompting method designed for LLMs to solve math problems with an arbitrary number of unknowns. We also show that our method performs competitively to state-of-the-art algorithms even for non-algebra problems in Appendix~\ref{sec:arithmetic_generalization}. 

A major challenge in applying the prompting method to multi-unknown problems is the scarcity of hand-crafted demonstrations. Traditional examples with a single unknown do not scale well to more complex, multi-unknown scenarios, necessitating automated demo generation. Furthermore, while LLMs can be guided by prompts to solve these systems of equations, they often require external tools due to their limited ability to independently solve and explicitly formulate these problems into a system of equations.

To overcome these limitations, we propose Formulate-and-Solve, a framework that incorporates a set of principles to instruct LLMs in generating demonstrations automatically. This framework empowers LLMs to translate problems into equations and subsequently utilize external tools to solve them. The overall pipeline is illustrated in Figure~\ref{fig:mwp_solver_fig} and we include the actual prompts used in each step in Appendix~\ref{tab:mwp_solver_instruction}.
% Our goal is to develop a generalized automatic problem solving framework for MWPs. The overall pipeline is presented in Algorithm 1. For MWP tasks, except for using only a natural language rationale to solve the problem, using a system of equations combined with natural language rationale excels in reasoning. Thus, the objective is to generate the system of equations that formalize MWPs based on a set of principles listed in Table X. Figure X shows an example used in the prompt that we created according to these principles. The full prompt is publicly available at [CODEBASE].

\begin{figure}[ht]
\centering
\includegraphics[width=0.48\textwidth]{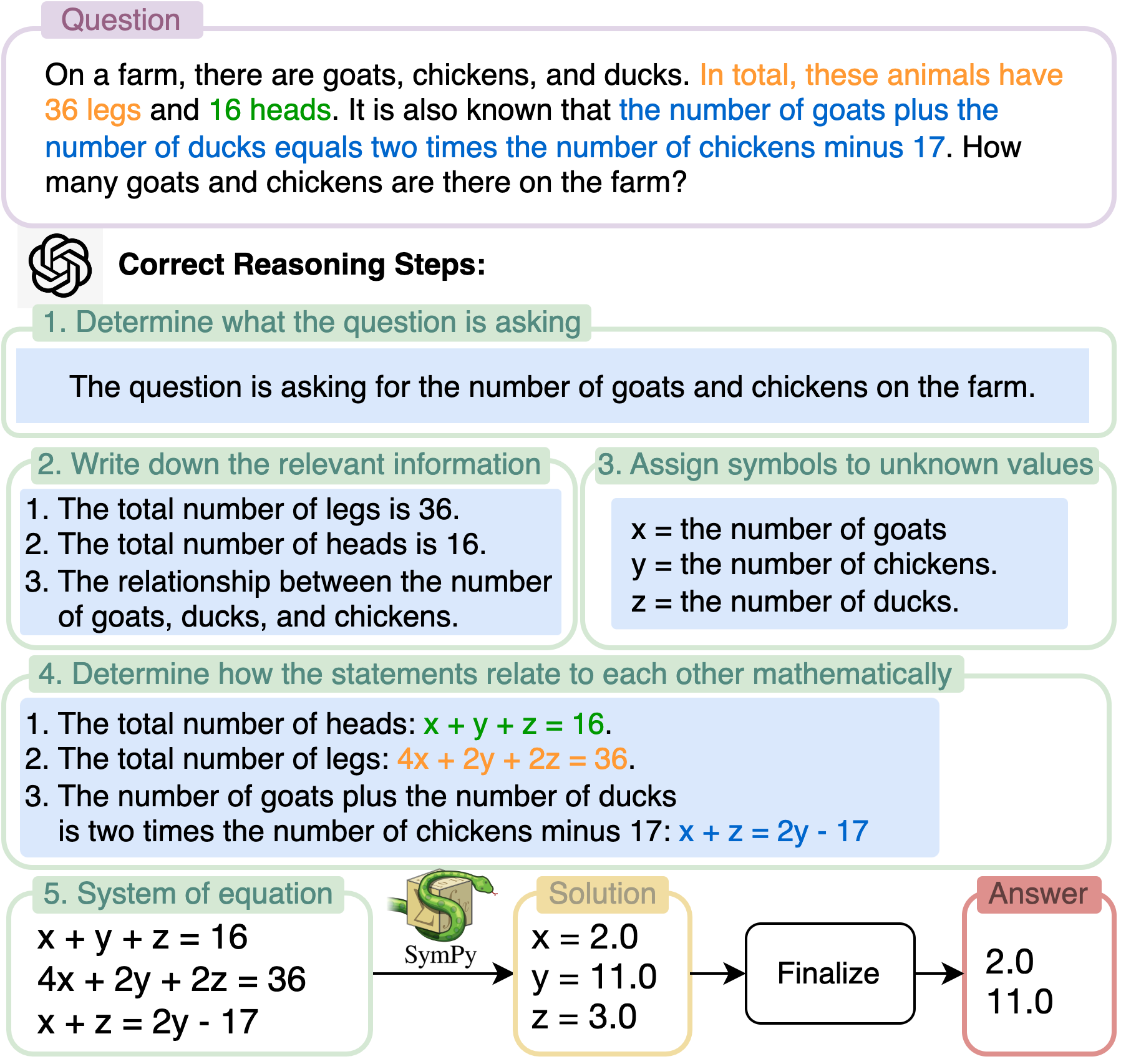}
\vspace{-4mm}
\captionsetup{font=small, width=0.9\linewidth}
\caption{The overview of Automatic Solver of Algebra Problems.}
\vspace{-6mm}
\label{fig:mwp_solver_fig}
\end{figure}

\paragraph{Automatic Generated Demonstrations.}
Conventional prompting methods require creating and evaluating human-written examples to guide LLMs in solving algebra problems, which is time-consuming.
Our approach leverages intuitive human-solving steps as instructions. Based on these instructions, the LLM iteratively generates its solution demonstrations (approximately five). To find the most effective demonstrations, we generate ten sets of demonstrations and assess their accuracy on twenty problems. The set with the highest accuracy is chosen as the best.
% Traditionally, prompting methods rely on the creation and evaluation of human-written exemplars to guide LLMs in solving algebra problems. This process can be time-consuming and labor-intensive. Our approach utilizes intuitive human-solving steps as instructions to equip LLMs to solve algebra problems. This instruction then serves as a foundation for the LLM to iteratively generate its own demonstrations (approximately five). Then, to efficiently identify effective demonstrations, we generate ten samples and assess their answer accuracy on twenty problems.

% Then, we run ten iterations for selecting the best accuracy of the generated prompt, and in each iteration we calculate the accuracy of the twenty problems.

\paragraph{Solving Strategy.}
%Although LLMs can address simple systems of equations, their accuracy decreases with increasing complexity of the system of equations or the presence of decimals. Thus, 
Our proposed method leverages the strengths of both LLMs and symbolic solvers.
We cooperate with the human-solving steps in the instruction to convert algebra problems into the corresponding systems of equations. Recognizing the limitations of LLMs for complex systems, an external symbolic solver (e.g., SymPy) is employed to solve the system of equations. In the cases of unsolvability or errors that occur in solving the system of equations, the finalization module relies solely on the original prompt and response. This strategy ensures adaptability in some scenarios where the response is not formatted. Furthermore, by incorporating the historical prompt and response within the finalization module, the approach facilitates the continuation or refinement of the solution response.

%% file: sections/06-Experiment.tex
\section{Experimental Results}

\subsection{Experimental Setting}
\label{Sec:exp_setting}

\paragraph{Dataset.}
Our experiments are conducted on five algebra problem sets, including existing widely-used ones (ALG514, DRAW-1K, AsDiv, HMWP) and the proposed BeyondX benchmark.
AsDiv consists of a wide range of math problems and we only take the algebra problem subset. Also, since HWMP is a Chinese dataset, we use GPT-4 to translate the problem into English. We find that while most translation results effectively convey the intended meaning.
The five datasets differ in size and complexity, as shown in Table~\ref{tab:dataset_statistics}. 
We also report the average number of unknowns in each dataset. Note that we split the proposed dataset into three subsets, correspond to problems with 3 (BeyondX\_3), 4 (BeyondX\_4), and 5 (BeyondX\_5) unknowns, while the problems in all other datasets have $\leq 2$ unknowns.

\input{tables/dataset_statistics_main_result}
\input{tables/GPT3.5_MU}
\input{tables/GPT3.5_main_table}

\paragraph{Models.}
For experiments in this section, we utilize GPT-3.5, GPT-4, and Gemini-Pro as representatives of general-purpose LLMs.
We also opt to exclude open-sourced LLMs, as they typically struggle with multi-unknown problems due to their limited capacity to process and follow long prompts. The versions of these models and system prompts we used for the experiments are listed in Appendix~\ref{sec:model_hyperparameter}.

\paragraph{Baselines.}
We compare Formulate-and-Solve with three types of prompting baselines: (1) \emph{Zero-shot.} We include Zero-shot-CoT~\cite{kojima2022large} and Plan-and-Solve (PS)~\cite{wang-etal-2023-plan} prompting. The former appends ``Let's think step by step'' to the prompt without any demo. The latter appends ``Let’s first understand the problem and devise a plan to solve the problem. Then, let’s carry out the plan and solve the problem step by step'' to the prompt without any demo. (2) \emph{Few-shot with manual demonstrations.} CoT~\cite{wei2022chain} creates eight hand-crafted natural language examples as demonstrations. PoT~\cite{gao2022pal} creates eight hand-crafted Python code examples as demonstrations and uses programming tools to get the final answer. EoT~\cite{liu2023plan} creates eight hand-crafted equation examples as demonstrations and uses symbolic solvers to obtain the final answer. Declarative (DR)~\cite{he2023solving} creates three hand-crafted examples with principles as demonstrations and uses symbolic solvers to obtain the final answer. (3) \emph{Few-shot with automatic demonstrations.} \cite{yasunaga2023large} proposed Analogical prompting (AG), designed to automatically guide LLMs to self-generate relevant examples as demonstrations before proceeding to solve the problem. We come up with another naive method:
selecting examples from the dataset and employing Zero-shot-CoT~\cite{kojima2022large} to generate examples as demonstrations. We refer to this method as Auto-Zero-shot-CoT (AZ).
%which is to select examples from the dataset and use Zero-shot-CoT~\cite{kojima2022large} to generate automatic examples as demonstrations, here we name Auto-Zero-shot-CoT (AZ).

% \input{latex/tables/dataset_statistics_main_result}
% \input{latex/tables/1variable}
% \input{latex/tables/GPT3.5_main_table}

\input{tables/ablation}

\subsection{Main Results}
To investigate which reasoning methods and models better solve multi-unknown problems, we summarize the performance of different prompting methods using GPT-3.5 in Table~\ref{tab:gpt3.5_mu_perf} and Tables~\ref{tab:gpt3.5_perf}. We also evaluate and compare various prompting methods with GPT-4 and Gemini-Pro in Appendix~\ref{sec:different_models}.

\paragraph{Comparison with various prompting baselines.}
Table~\ref{tab:gpt3.5_mu_perf} and Figure~\ref{fig:different_method_alg514} present the accuracy comparison of our method with existing approaches on datasets containing 1 to 5 unknowns. We combine ALG514 and DRAW-1K into MU\_1 and MU\_2 for problems with 1 and 2 unknowns, respectively, and use BeyondX for problems with 3, 4, or 5 unknowns. Figure \ref{fig:different_method_alg514} illustrates that only our method maintains reasonable accuracy when the number of unknowns exceeds two, while other methods experience a significant performance decline.  For problems with five unknowns, our method achieves a 66.7\% accuracy, while the best alternative achieves only 34.2\%. 

This improvement is attributed to our instructional approach and automatic demonstrations, which effectively address general algebra problems. Furthermore, datasets with multiple unknowns typically involve longer questions and necessitate the construction of more equations for a solution. We find that using lengthy natural language reasoning steps can easily introduce operational and calculation errors. In contrast, our method guides LLMs to generate a system of equations as an intermediate reasoning step, making it less prone to mistakes during equation formulation. As a result, our method maintains high accuracy when using a symbolic solver to solve the equations. We also observe that leveraging external tools, such as programming or symbolic solvers, to tackle algebra problems generally yields better performance than directly obtaining the final answer from the model in the presence of multiple unknowns.

Next, we verify whether Formulate-and-Solve maintains its effectiveness for problems with one or two unknowns. For this, we compare it with baselines on commonly used algebra datasets containing one or two unknowns, with the results presented in Table \ref{tab:gpt3.5_perf}.
The results demonstrate that Formulate-and-Solve again achieves the best performance. 
Compared with other automatic few-shot methods such as AZ, the performance gap is considerably smaller (5.2\% on average) for one or two unknowns than for multiple unknowns (48.4\% on average). This is likely because it is easier to generate a high-quality demo for problems with one or two unknowns. 
Also, we cannot see a big difference between the zero-shot and few-shot CoT in this experiment since the manual few-shot demonstrations that are commonly used in previous work are beneficial for solving arithmetic problems, not algebra problems. 
% Furthermore, we conduct experiments on arithmetic datasets to evaluate the generalizability of Fomulate-and-Solve in Appendix~\ref{sec:arithmetic_generalization}.

% \begin{figure}[h]
% \includegraphics[width=\columnwidth]{latex/figures/multiple_var_comparison.png}
%    \caption{Comparison of different number of variables between different methods.}
% \label{fig:multiple_var_comparison_fig}
% \end{figure}

%% file: tables/dataset_statistics_main_result.tex
\begin{table}[ht]
\centering
\small
\resizebox{7.5cm}{!}{%
\begin{tabular}{lccc}
\thickhline
\textbf{Dataset} & \textbf{Avg \#unknowns} & \textbf{Avg \#words} & \textbf{Size} \\ \hline
ALG514  & 1.8            & 41.4        & 514  \\
DRAW-1K  & 1.7            & 35.8        & 200  \\
AsDiv   & 1.6            & 27.9        & 154  \\
HMWP    & 1.3            & 77.3        & 548  \\
BeyondX\_3   & 3              & 65.1        & 194  \\
BeyondX\_4   & 4              & 94.5        & 158  \\
BeyondX\_5   & 5              & 118.8       & 112  \\ \thickhline
\end{tabular}}
\vspace{-2mm}
\captionsetup{font=small, width=0.9\linewidth}
\caption{Statistics of existing algebra dataset.}
\vspace{-6mm}
\label{tab:dataset_statistics}
\end{table}

%% file: tables/GPT3.5_MU.tex
\begin{table*}[ht]
\centering
\small
\resizebox{14cm}{!}{
\begin{tabular}{l|cc|cccc|ccc}
% \begin{tabularx}{\linewidth}{lXX|XXXX|XXX}
\thickhline
Setting                                              & \multicolumn{2}{c|}{Zero-shot} & \multicolumn{4}{c|}{Few-shot (Manual)}  & \multicolumn{3}{c}{Few-shot (Automatic)} \\ \hline
Method                                               & CoT       & PS    & CoT    & PoT$^\dagger$    & EoT$^\dagger$    & DR$^\dagger$ & AG & AZ & Ours$^\dagger$   \\ \hline
\multicolumn{1}{l|}{MU\_1}                          & 79.3\%    & 77.2\%            & 82.8\% & 80.0\% & 25.5\% & 51.7\%      & 33.8\%     & \underline{90.3\%}              & \textbf{89.7\%} \\
\multicolumn{1}{l|}{MU\_2}                         & 46.2\%    & 40.8\%            & 52.7\% & 77.3\% & 26.7\% & 67.0\%      & 33.9\%     & \underline{87.9\%}             & \textbf{93.7\%} \\
\multicolumn{1}{l|}{BeyondX\_3}                          & 11.3\% & 9.8\%  & 12.4\% & 52.1\% & 3.6\%  & \underline{55.2\%} & 10.3\% & 0.5\%  & \textbf{93.8\%} \\
\multicolumn{1}{l|}{BeyondX\_4}                          & 6.3\%  & 8.9\%  & 3.2\%  & \underline{32.3\%} & 0.0\%  & 21.5\% & 10.1\% & 2.5\%  & \textbf{84.9\%} \\
\multicolumn{1}{l|}{BeyondX\_5}                          & 6.2\%  & 5.4\%  & 2.7\%  & \underline{36.6\%} & 0.0\%  & 5.4\%  & 9.8\%  & 1.8\%  & \textbf{71.4\%} \\
\rowcolor[HTML]{DEDEDE}
\multicolumn{1}{l|}{Average} & 29.9\% & 28.4\% & 30.7\% & \underline{55.7\%} & 11.2\% & 40.1\% & 19.6\% & 36.6\% & \textbf{86.7\%} \\ \thickhline
\end{tabular}}
\vspace{-2mm}
\captionsetup{font=small, width=0.9\linewidth}
\caption{Experiment results across various unknowns using GPT-3.5. $^\dagger$ means whether the method uses external tools or not.}
\vspace{-3mm}
\label{tab:gpt3.5_mu_perf}
\end{table*}

%% file: tables/GPT3.5_main_table.tex
\begin{table*}[htb!]
\centering
\small
\resizebox{14cm}{!}{%
\begin{tabular}{l|cc|cccc|ccc}
% \begin{tabularx}{\linewidth}{lXX|XXXX|XXX}
\thickhline
Setting                                              & \multicolumn{2}{c|}{Zero-shot} & \multicolumn{4}{c|}{Few-shot (Manual)}  & \multicolumn{3}{c}{Few-shot (Automatic)} \\ \hline
Method                                               & CoT       & PS    & CoT    & PoT$^\dagger$    & EoT$^\dagger$    & DR$^\dagger$ & AG & AZ & Ours$^\dagger$   \\ \hline
\multicolumn{10}{c}{\textit{Single}}                                                                                                                                              \\ \hline
\multicolumn{1}{l|}{ALG514}                          & 91.2\%    & 85.7\%            & 91.2\% & \underline{93.4\%} & 31.9\% & 59.3\%      & 38.5\%     & \textbf{93.4\%}             & 92.3\% \\
\multicolumn{1}{l|}{DRAW-1K}                         & 59.3\%    & 63.0\%            & 68.5\% & 57.4\% & 14.8\% & 38.9\%      & 25.9\%     & \underline{85.2\%}             & \textbf{85.2\%} \\
\multicolumn{1}{l|}{AsDiv}                           & 68.3\%    & 71.4\%            & 71.4\% & 75.0\% & 26.8\% & 53.5\%      & 54.0\%     & \underline{76.1\%}             & \textbf{81.7\%} \\
\multicolumn{1}{l|}{HMWP}                            & 21.5\%    & 14.6\%            & 22.0\% & 25.2\% & 4.2\%  & 10.3\%      & 21.0\%     & \underline{31.0\%}             & \textbf{37.7\%} \\
\rowcolor[HTML]{DEDEDE}
\multicolumn{1}{l|}{Average} & 60.1\%    & 58.7\%            & 63.3\% & 62.8\% & 19.4\% & 40.5\%      & 34.8\%     & \underline{71.4\%}             & \textbf{74.2\%} \\ \hline
\multicolumn{10}{c}{\textit{Double}}                                                                                                                                              \\ \hline
\multicolumn{1}{l|}{ALG514}                          & 47.0\%    & 40.7\%            & 53.4\% & 81.8\% & 29.6\% & 68.8\%      & 34.5\%     & \underline{90.5\%}             & \textbf{96.5\%} \\
\multicolumn{1}{l|}{DRAW-1K}                          & 43.8\%    & 41.1\%            & 50.7\% & 64.4\% & 18.5\% & 61.6\%      & 32.2\%     & \underline{80.1\%}             & \textbf{85.6\%} \\
\multicolumn{1}{l|}{AsDiv}                           & 55.0\%    & 49.5\%            & 37.4\% & 11.0\% & 15.7\% & 59.0\%      & 31.9\%     & \underline{79.5\%}             & \textbf{80.7\%} \\
\multicolumn{1}{l|}{HMWP}                            & 16.4\%    & 9.9\%             & 18.7\% & 39.8\% & 9.9\%  & 26.3\%      & 28.1\%     & \underline{50.3\%}             & \textbf{58.5\%} \\
\rowcolor[HTML]{DEDEDE}
\multicolumn{1}{l|}{Average} & 40.6\%    & 35.3\%            & 40.0\% & 49.2\% & 18.4\% & 53.9\%      & 31.7\%     & \underline{75.1\%}             & \textbf{80.3\%} \\ \thickhline
% \end{tabularx}}
\end{tabular}}
\vspace{-2mm}
\captionsetup{font=small, width=0.9\linewidth}
\caption{Experiment results across various algebra problem datasets which include single and double unknown using GPT-3.5. $^\dagger$ means whether the method uses external tools or not.}
\label{tab:gpt3.5_perf}
\vspace{-6mm}
\end{table*}

%% file: tables/ablation.tex
\begin{table*}[h]
\centering
\small
\resizebox{14cm}{!}{%
\begin{tabular}{ccccccccc}
\thickhline
\textbf{Instruction} & \textbf{Demos} & \textbf{Solver} & \textbf{Equation} & \textbf{U1}      & \textbf{U2}      & \textbf{U3}      & \textbf{U4}      & \textbf{U5}      \\ \hline
\xmark           & \xmark          & \cmark      & \cmark        & 77.8\% & 74.0\% & 60.0\% & 33.8\% & 18.3\% \\
\xmark           & \cmark          & \cmark      & \cmark        & \underline{79.6\%} & 77.4\% & 79.0\% & 56.3\% & 48.3\% \\
\cmark           & \xmark          & \cmark      & \cmark        & 64.1\% & \underline{81.2\%} & \underline{82.0\%} & \underline{68.3\%} & \underline{60.0\%} \\
\cmark           & \cmark          & \xmark      & \cmark        & 59.3\% & 73.3\% & 43.0\% & 16.8\% & 15.0\% \\
\cmark           & \cmark          & \cmark      & \cmark        & \textbf{85.2\%} & \textbf{87.6\%} & \textbf{90.0\%} & \textbf{81.3\%} & \textbf{66.7\%} \\ \thickhline
\end{tabular}}
\captionsetup{font=small, width=0.9\linewidth}
\caption{Ablation experiment results across various number of unknowns using GPT-3.5. Our method achieves the highest accuracy among all.}
\vspace{-6mm}
\label{tab:ablation}
\end{table*}

%% file: sections/07-Discussion.tex
\section{Discussion and Analysis}

\paragraph{Ablation study.}
We analyze the significance of each component within in Formulate-and-Solve through an ablation study. We assess five variations: (1) Use only a system of equations as a rationale for reasoning. (2) Remove the instruction before demonstrations. (3) Remove demonstrations after the instruction. (4) Use an LLM instead of a symbolic solver to solve a system of equations. (5) Formulate-and-Solve. We randomly select 60 problems for each unknown from ALG514, DRAW-1K, and BeyondX to evaluate each variation. The results are provided in Table~\ref{tab:ablation}.

We observe that performance decreases significantly when instruction or demonstration is removed, highlighting its role in guiding the LLM. Interestingly, instruction has a greater impact than demonstration. Replacing the symbolic solver with an LLM also leads to a large decrease in accuracy. These findings confirm that all elements in Formulate-and-Solve contribute significantly to solving multi-unknown problems.
% When the instruction before demonstrations is removed (second row), the performance drops notably, indicating the importance of instruction in guiding the generation of demonstrations. Similarly, removing demonstrations after instruction (third row) results in a significant decrease in performance, highlighting the need to provide examples to facilitate understanding. Interestingly, prepending the instruction leads to a greater influence than demonstrations. Using an LLM instead of a symbolic solver (fourth row) also leads to a considerable decrease in accuracy, demonstrating the effectiveness of a symbolic solver in accurately solving the system of equations.
% In conclusion, the ablation study confirms that each component of our proposed method is crucial to achieving optimal performance. The combination of instruction, demonstrations, a system of equations, and a symbolic solver significantly enhances the problem-solving capabilities in multiple unknown problems.
\input{tables/error_analysis}

\paragraph{Error analysis.}
We delve deeper into the primary challenges that LLMs encounter when solving multi-unknown problems using Formulate-and-Solve. To gain a quantitative understanding of model failures, we conduct an error analysis in Formulate-and-Solve with GPT-3.5 on BeyondX. We collect all instances where predictions were incorrect and annotate the main reasons for these mispredictions. The error types include: (E1) generating too few or too many equations, (E2) producing the correct number of equations but with incorrect content, and (E3) generating responses in the wrong format, preventing the extraction of the equation system. 
%generating a correct system of equations, but the original problem has an infinite number of solutions, and 

As illustrated in Table~\ref{tab:error_analysis_table}, the most common error is E2 (incorrect equation). This indicates that current LLMs equipped with prompting methods still struggle to accurately formulate multi-unknown equations in some cases. Besides, 37.7\% of the errors occur due to the wrong format of the response, and 17.4\% of the errors arise when LLMs fail to align relevant information correctly with the equations, resulting in either too few or too many equations. The detailed qualitative analysis of the error examples is in Appendix~\ref{sec:qualitative_error_cases}.

% We also show every type of error case that GPT-3.5 cannot answer correctly in Table~\ref{tab:error_example_1}, Table~\ref{tab:error_example_2}, and Table~\ref{tab:error_example_3}. From E1, the system of equations is missing an equation about the relation "Total number of cars: 20". From E2, the first equation is wrong since the relation is "Total sum of the average miles per gallon obtained by the three cars is 75", which means "a + b + c = 75". From E3, since the first equation equals the third equation twice, the solution of this system of equations is infinite. From E4, since the response format is different from the demonstration, we cannot extract the system of equations from the response.

%% file: tables/error_analysis.tex
\begin{table}[ht!]
\centering
\small
\resizebox{7cm}{!}{%
\begin{tabular}{lc}
\thickhline
\textbf{Error Reason} & \textbf{Percentage (\%)} \\ \hline
(E1) Fewer or More Equations & 17.4\% (12/69) \\ 
(E2) Incorrect Equation & 44.9\% (31/69) \\ 
(E3) Incorrect Extraction Form & 37.7\% (26/69) \\ \thickhline
\end{tabular}}
\vspace{-2mm}
\captionsetup{font=small, width=0.9\linewidth}
\caption{Statistics of Error Analysis under GPT-3.5.}
\vspace{-6mm}
\label{tab:error_analysis_table}
\end{table}

%% file: sections/08-Conclusion.tex
\section{Conclusion}
We introduce BeyondX, the first benchmark for evaluating LLMs on multi-unknown problems. Our analysis reveals a significant performance drop in LLMs and existing prompting methods when faced with such problems. To address this, we propose Formulate-and-Solve, a novel prompting method that leverages instruction, automatic demonstrations and a system of equations. Experiments demonstrate the effectiveness of Formulate-and-Solve in tackling multi-unknown problems.

%% file: sections/09-Limitations.tex
\section*{Limitations}

\subsection*{Scope of Benchmark.} Although our automatic generation method can decrease the labor-intensive data collection process, our method still needs to be expanded from high-quality problems with low unknowns. Besides, we figure that some types or topics of the problems cannot be extended to multiple unknown problems. And, our benchmark is limited to English questions and data. We look forward to future benchmarks on a broader domain or modality and other languages.

\subsection*{Models and Reasoning Methods.} Although we experiment with many representative models and reasoning methods in this paper, we acknowledge that this does not cover all models and frameworks. Besides, we acknowledge that our approach falls short on more straightforward arithmetic datasets since our method is more suitable for algebra datasets. Further research is required to explore new problem-solving methods for general math reasoning tasks, including different modalities.

%% file: sections/10-Appendix.tex
\clearpage
\appendix

% \section{Failure Cases of Intuitive Generation Method}

% \ref{tab:direct_mwp_generation} show some failure cases using intuitive ways to prompt LLMs to generate multiple unknown problems. Here, we use prompt GPT-4 to generate three-unknown problems and its corresponding system of equations.

% \section{Failure Cases of Proposed Generation Method}
\input{tables/dataset_overview}
\input{tables/arithmetic_experiment}
\input{tables/open_source_model}
\input{tables/hyperparameters}
\input{tables/dataset_error_rate}
\input{tables/different_shot}

\begin{algorithm}[h]
\caption{Formulate-and-Solve Reasoning Algorithm}
\label{alg:mwp_solver}
\begin{algorithmic}[1]
\Require question $Q$, instruction $I$, reasoning module $R$, symbolic solver $S$, finalize module $F$
\Function{Auto Demo}{$I$, $Question$, $K$}
\While{$K \neq 0$}
\State $D \gets R(I+D+Question)$ 
\State $K \gets K - 1$
\EndWhile
\State \Return $D$ \Comment{$D$ is a demo}
\EndFunction
\State $D \gets Auto Demo()$
\State $p \gets I + D + Q$ \Comment{$p$ is a input prompt}
\State $eq \gets R(p)$
\If{$S(eq)$} \Comment{Equation System is solvable}
    \State $ans \gets F(Q + eq + S(eq))$
\Else
    \State $ans \gets F(Q + eq)$
\EndIf
\State \Return $ans$ \Comment{Return the Answer}
\end{algorithmic}
\end{algorithm}

\section{Further Experiment}

\subsection{Generalization to common arithmetic datasets}
\label{sec:arithmetic_generalization}

We analyze the generalizability of the Formulate-and-Solve framework to other common arithmetic datasets, such as GSM8K, SVAMP, AddSub, SingleEq, and MultiArith where some problems can be seen as single unknown problems on GPT-3.5. From Table ~\ref{tab:gpt3.5_arithmetic_perf}, we can see that our method can still perform a comparable performance to other existing custom and manual prompting methods for arithmetic tasks since these datasets are much easier than multiple unknown datasets and allow only minimal room for improvement. Besides, our method is more generalized not only to single unknown problems but also can deal with multiple unknown problems automatically.

\subsection{Experiments different mathematical models on BeyondX}

\begin{figure}[h]
\centering
\includegraphics[width=0.9\linewidth]{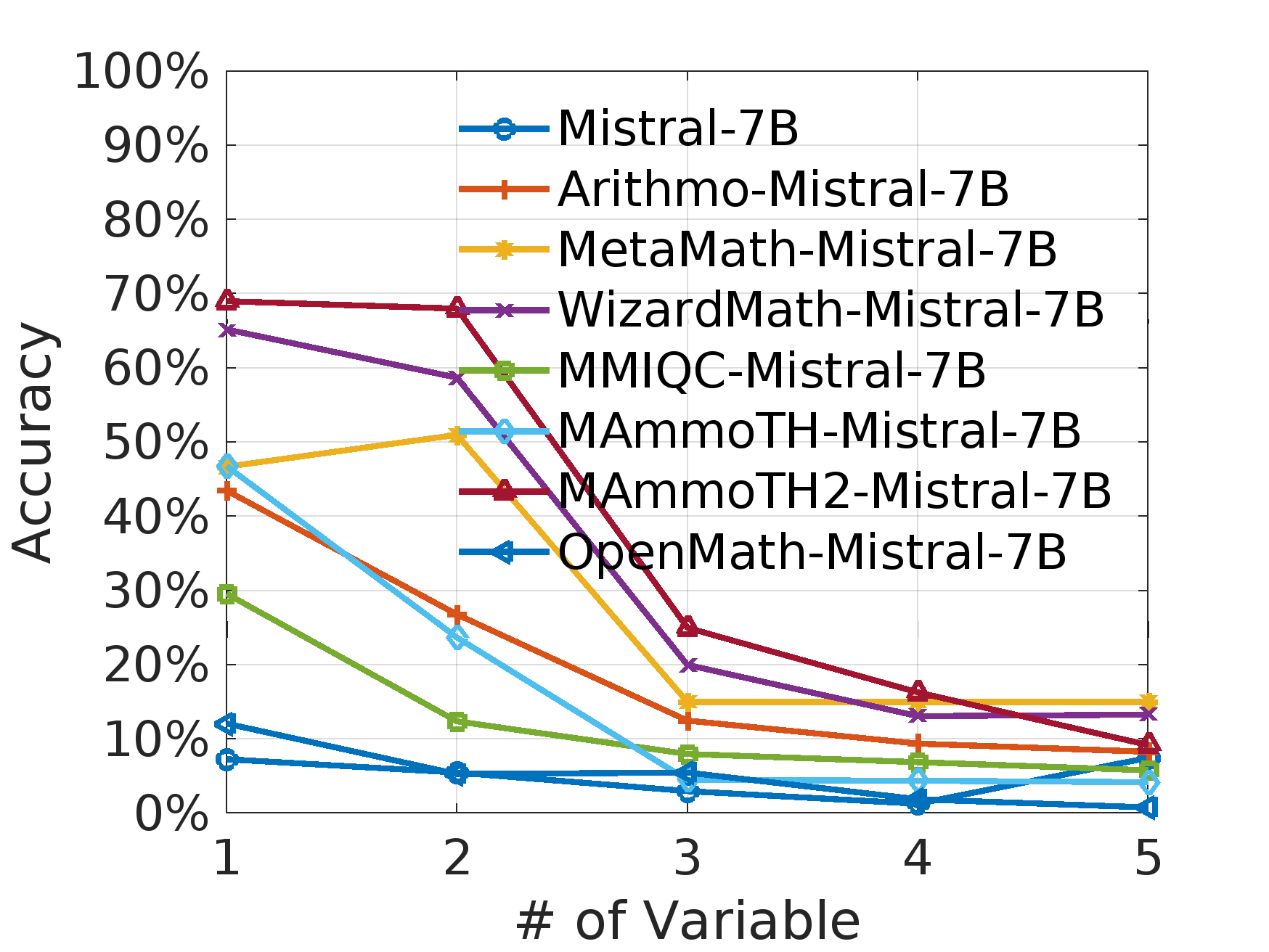}
\captionsetup{font=small, width=0.9\linewidth}
\caption{The performance of different existing open-source models.}
\label{fig:open_source_appendix}
\end{figure}

\begin{figure*}[h]
 \centering
  \begin{subfigure}{0.477\textwidth}
    \includegraphics[width=\linewidth]{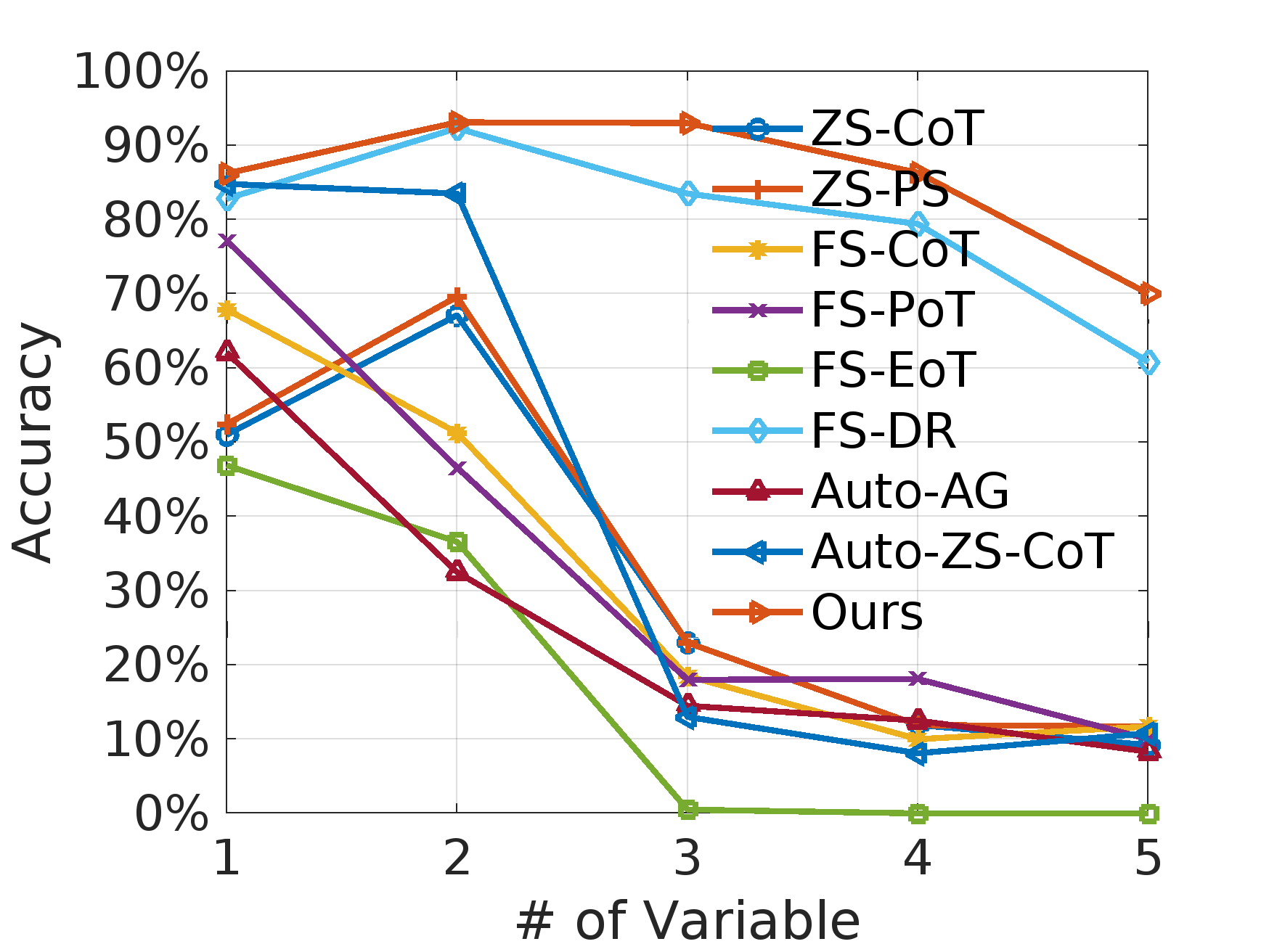}
    \captionsetup{font=small, width=0.8\linewidth}
    \caption{Different prompting methods performance of Gemini-Pro.}
    \label{fig:different_model_Gemini}
  \end{subfigure}
  \hfill % This command adds a horizontal space between the subfigures
  \begin{subfigure}{0.477\textwidth}
    \includegraphics[width=\linewidth]{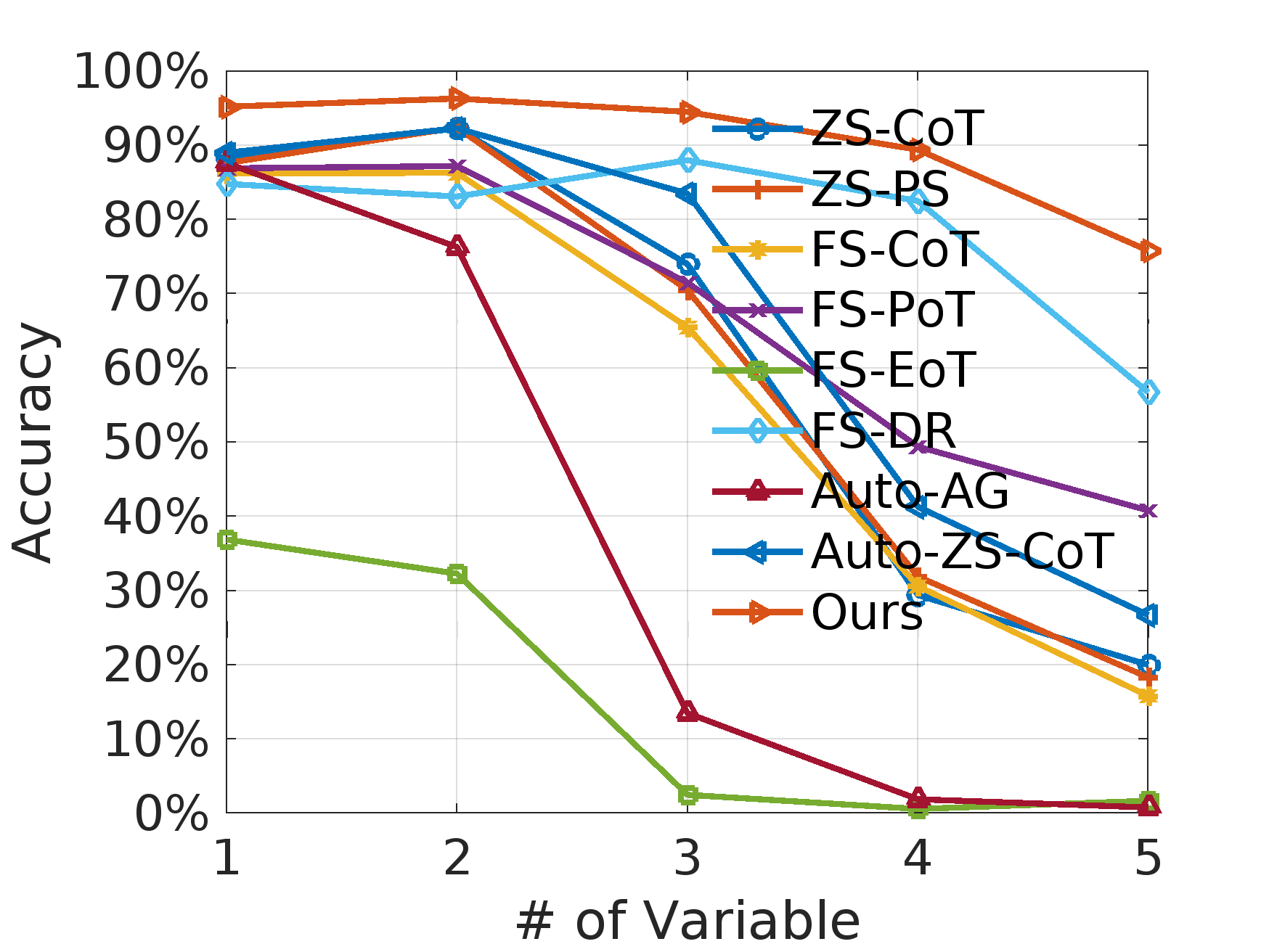}
    \captionsetup{font=small, width=0.8\linewidth}
    \caption{Different prompting methods performance of GPT-4.}
    \label{fig:different_model_GPT4}
  \end{subfigure}
  \captionsetup{font=small, width=0.9\linewidth}
 \caption{The performance on different numbers of unknown.}
 \label{fig:different_model_appendix}
\end{figure*}

\input{tables/model_MU}
\input{tables/model_main_table}
\input{tables/MWP_generation_prompt}
\input{tables/MWP_solver_prompt}

\begin{figure*}[h]
 \centering
  \begin{subfigure}{0.477\textwidth}
    \includegraphics[width=\linewidth]{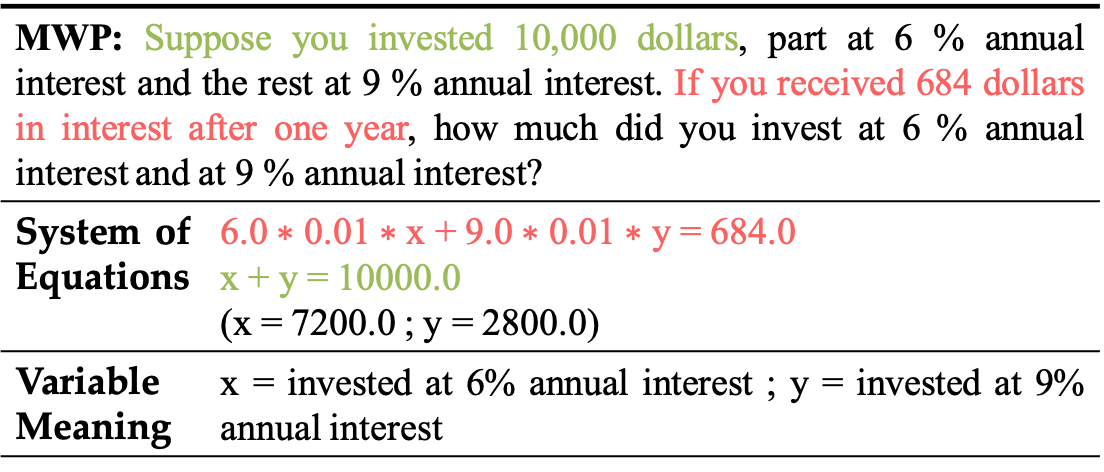}
    \captionsetup{font=small, width=0.8\linewidth}
    \caption{Two-unknown (source problem)}
    \label{fig:mwp_2}
  \end{subfigure}
  \hfill % This command adds a horizontal space between the subfigures
  \begin{subfigure}{0.477\textwidth}
    \includegraphics[width=\linewidth]{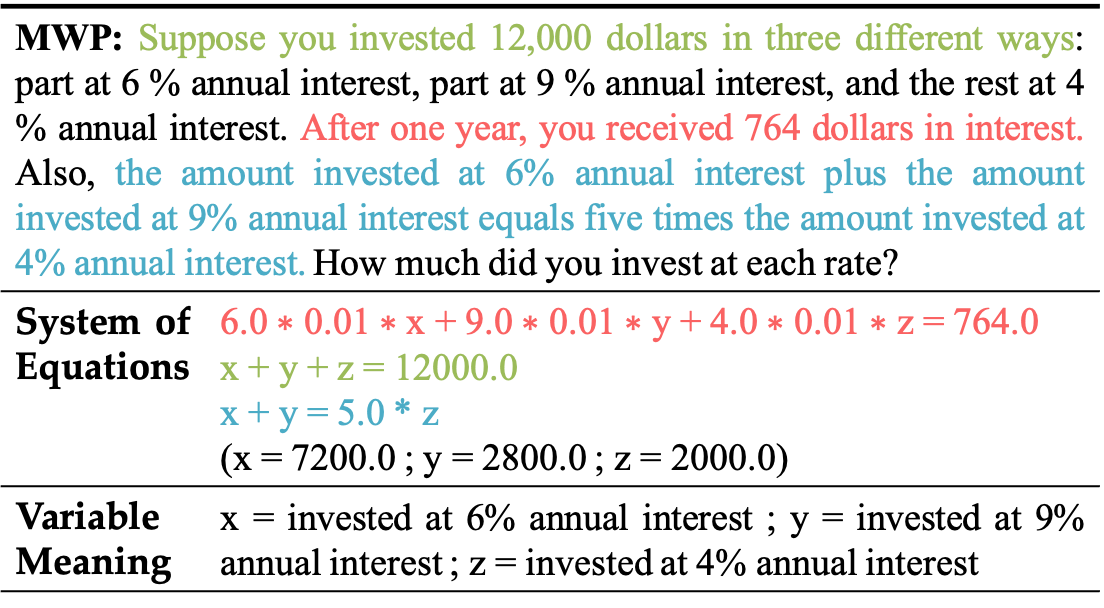}
    \captionsetup{font=small, width=0.8\linewidth}
    \caption{Three-unknown}
    \label{fig:mwp_3}
  \end{subfigure}
  \begin{subfigure}{0.477\textwidth}
    \includegraphics[width=\linewidth]{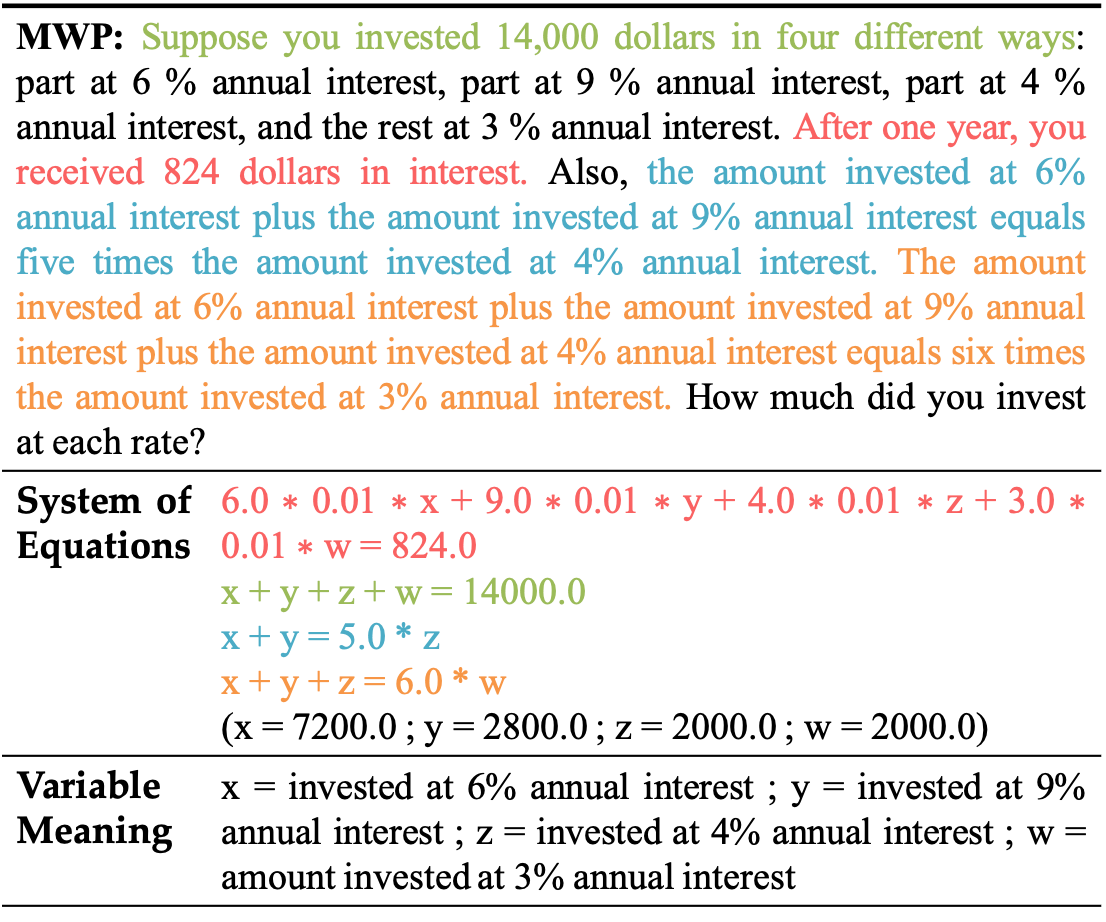}
    \captionsetup{font=small, width=0.8\linewidth}
    \caption{Four-unknown}
    \label{fig:mwp_4}
  \end{subfigure}
  \hfill
  \begin{subfigure}{0.477\textwidth}
    \includegraphics[width=\linewidth]{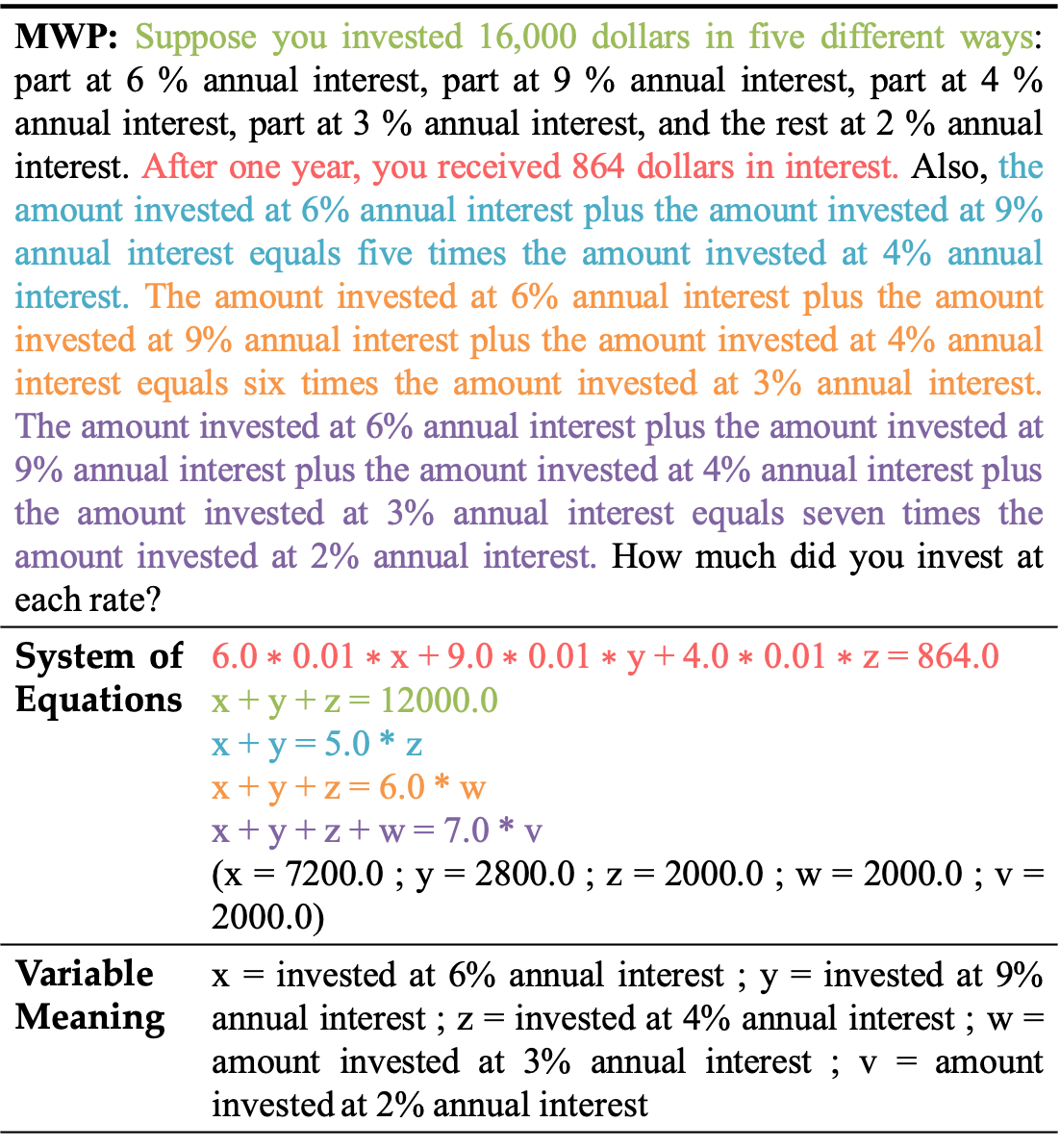}
    \captionsetup{font=small, width=0.8\linewidth}
    \caption{Five-unknown}
    \label{fig:mwp_5}
  \end{subfigure}
\captionsetup{font=small, width=0.9\linewidth}
 \caption{Example problems with different numbers of unknown generated from the same source problem.}
 \label{fig:examples_mwp}
\end{figure*}

\input{tables/direct_prompt_for_mwp_generation}
\input{tables/equation_aware_mwp_generation}
\input{tables/error_cases_mwp_generation}

We further evaluate seven different existing mathematical models fine-tuned on Mistral-7B on \textbf{BeyondX} under Zero-shot-CoT setting. As shown in Figure~\ref{fig:open_source_appendix}, the results indicate that these open-source LLMs are still struggling with more complex mathematical reasoning tasks in multiple unknown problems. There is still a significant amount of effort during pretraining or supervised fine-tuning to instill enough multiple unknown knowledge and the way of solving multiple unknown system of equations into the models’ parameters to close the gap.

\subsection{Experiments different models on Formulate-and-Solve}
\label{sec:different_models}
We further assess the performance of Formulate-and-Solve across various base models, such as GPT-4 and Gemini-Pro. The results of the general algebra dataset and the multi-unknown dataset are shown in Table~\ref{tab:model_main_perf} and Table~\ref{tab:model_mu_perf}, and we also illustrate the performance curve of GPT-4 and Gemini-Pro on different numbers of unknown in Figure~\ref{fig:different_model_Gemini} and Figure~\ref{fig:different_model_GPT4}. The findings remain as GPT-3.5 and our method outperforms a large gap among other methods. Additionally, we observe that the performance of the Gemini-Pro generally falls between that of GPT-3.5 and GPT-4 across various datasets and prompting methods.

\subsection{Experiments different shots of demonstrations on Formulate-and-Solve}
We analyze the effect of varying the number of automatic generated exemplars (K) in our approach on GPT-3.5. Here, we show three variations with K = 3, 5, and 8. In Table~\ref{tab:different_shot}, we observe that LLM demonstrates consistent performance under single or double unknown in different datasets. When K is bigger, on average, performance improves.

\section{Full Instruction}
In this section, we show the full instructions in Section~\ref{sec:generation} and Section~\ref{sec:solver}. 

\subsection{Automatic Generation of Multiple Unknown Algebra Problems}

We can see the full instructions in Table~\ref{tab:mwp_generation_instruction}.

\subsection{Automatic Solver of Algebra Problems}
We can see the full instructions in Table~\ref{tab:mwp_solver_instruction}.

\section{Detail Studies of Automatic Generation of Multi-Unknown Algebra Problems}

\subsection{Construction Steps}
Starting from the 2 unknown problems in our seed dataset ALG514 and DRAW-1K, we use Zero-shot prompting with GPT-4 (\texttt{gpt-4-0613}) to generate an initial demonstration using the instruction in Table~\ref{tab:mwp_generation_instruction}, which is then manually refined. The LLM then iteratively creates additional demonstrations (approximately five) based on the problem, the system of equations, and the existing demonstrations. Combining this information, the LLM generates a new problem with N+1 unknowns and its corresponding system of equations. Finally, we use GPT-4 to solve these newly generated problems, discarding low-quality ones (where GPT-4 provides incorrect answers) for the next round.

\subsection{Quality Validation}
We recruit 12 raters to validate whether the generated problems are reasonable and whether they are consistent with the generated system of equations. %They record the index of reasonable generated problems and the number of problems they saw in each unknown subset. 
We show the ratio of problems that are marked as ``unreasonable'' by raters
%We show the unreasonable problem rate (\# Unreasonable Problem/\# Total
%Human Seen Problem) for each subset 
in Table~\ref{tab:error_rate_dataset}. 
To understand why LLMs struggle with our instruction, we analyzed unreasonable problems in Table~\ref{tab:error_cases_mwp_generation}. Our findings reveal several limitations. First, LLMs cannot directly derive complex constant meanings requiring decomposition (Case 1). Second, unclear instructions lead to repetitive equations (Case 2). Third, introducing new variables might not be effective for all problems (e.g., river rate and distance, Case 3). Finally, LLMs may generate inconsistent numerical values within problems (Case 4).
% To find the insightful reason why LLM cannot be instructed successfully, we perform an error analysis for unreasonable problems and provide some cases in Table~\ref{tab:error_cases_mwp_generation}. Our analyses are summarized as follows. From Case 1, if the constant in the equation contains complex meaning and needs to be decomposed into different statements, LLM cannot derive directly. From Case 2, the absence of clear instructions during equation generation can lead to repetitive equations from LLMs. From Case 3, for certain problems like river rate and distance traveled, directly introducing new variables might not be the most effective approach. From Case 4, LLMs may struggle to generate consistent numerical values within a problem for a given system of equations.
% {\color{red}(cho: maybe briefly talk about the results here (e.g., we verified that the generated set is of high quality? Or we need to use human feedback to correct/remove some of the problems. In that case, what's the final statistics of the dataset? )}
% In Section 6, we review the incorrect examples that LLMs cannot expand.

% give more details why we need to analyze unreasonable problems and move to appendix

\subsection{Full Examples}
In Figure~\ref{fig:examples_mwp}, we show the full examples of our proposed generation method under each unknown.

\section{Detail Studies of Automatic Solver of Algebra Problems}

\subsection{Model Hyperparameters}
\label{sec:model_hyperparameter}
The hyperparameters for the experiments for studying Formulate-and-Solve and other prompting methods are set to their default values to ensure consistency in our experiment. Table~\ref{tab:hyperparameter} details the specific generation parameters for the various LLMs we evaluate.

\subsection{Overall Pipeline}
We describe the overall pipeline of Formulate-and-Solve in Algorithm~\ref{alg:mwp_solver}.

\subsection{Qualitative Analysis of Error Cases}
\label{sec:qualitative_error_cases}
We show every type of error case that GPT-3.5 cannot answer correctly in Table~\ref{tab:error_example_1}. From E1, the system of equations is missing an equation about the relation "Total number of cars: 20". From E2, the first equation is wrong since the relation is "Total sum of the average miles per gallon obtained by the three cars is 75", which means "a + b + c = 75". From E3, since the response format is different from the demonstration, we cannot extract the system of equations from the response.

\subsection{Full Examples}
In Table~\ref{tab:full_example_1} and Table~\ref{tab:full_example_3}, we show the input and output examples for each unknown using Formulate-and-Solve in GPT-3.5. 

\input{tables/error_cases_mwp_solver}
\input{tables/full_example}

%% file: tables/dataset_overview.tex
\begin{table*}[t]
\centering
\small
\resizebox{15cm}{!}{%
\setlength{\tabcolsep}{1.5mm}{
\begin{tabular}{llllll}
\thickhline
\textbf{Dataset}     & \textbf{Rationale}        & \textbf{Size}   & \textbf{\# of Variables} & \textbf{Source}         & \textbf{Domain}             \\ \hline 
SingleEQ    & Equation         & 508    & 1                          & Internet       & Arithmetic         \\ 
MAWPS       & Equation         & 3320   & 1                          & Internet       & Arithmetic         \\ 
AllArith    & Equation         & 831    & 1                          & Internet       & Arithmetic         \\ \hline
Dolphin18K  & Equation         & 18460  & 1                          & Internet       & Arithmetic/Alegbra \\ 
Math23K     & Equation         & 23162  & 1                          & Internet       & Arithmetic/Alegbra \\ 
SVAMP       & Equation         & 1000   & 1                          & Internet       & Arithmetic/Alegbra \\ 
GSM8K       & Natural Language & 8792   & 1                          & Annotated      & Arithmetic/Alegbra \\ 
AQuA        & Natural Language & 100000 & 1                          & GMAT/GRE       & Arithmetic/Alegbra \\ 
MATHQA      & Natural Language & 37297  & 1                          & GMAT/GRE       & Arithmetic/Alegbra \\ \hline
ASDiv       & Equation         & 2305   &  1-2                 & Internet       & Arithmetic/Alegbra \\ 
ALG514      & Equation         & 514    & 1-2                 & Internet       & Alegbra            \\ 
DRAW-1K      & Equation         & 1000   & 1-2                 & Internet       & Alegbra            \\ 
HMWP        & Equation         & 5470   & 1-2                 & Internet       & Alegbra            \\ \hline
\textbf{BeyondX} & \textbf{Equation}         & \textbf{480}    &  \textbf{1-5}                          &  \textbf{LLMs Generated} & \textbf{Alegbra} \\
\thickhline
\end{tabular}}}
\vspace{-2mm}
\captionsetup{font=small, width=0.9\linewidth}
\caption{List of existing math dataset.}
\vspace{-2mm}
\label{tab:dataset}
\end{table*}

%% file: tables/arithmetic_experiment.tex
\begin{table*}[h]
\centering
\small
\resizebox{15cm}{!}{
\begin{tabular}{lcc|cccc|ccc}
% \begin{tabularx}{\linewidth}{XXXXXXXXXX}
\thickhline
Setting                                              & \multicolumn{2}{c|}{Zero-shot} & \multicolumn{4}{c|}{Few-shot (Manual)}  & \multicolumn{3}{c}{Few-shot (Automatic)} \\ \hline
Method                                               & CoT       & PS    & CoT    & PoT$^\dagger$    & EoT$^\dagger$    & DR$^\dagger$ & AG & CoT & Ours$^\dagger$   \\ \hline
\multicolumn{1}{l}{MultiArith}                          & 94.3\%    & 95.3\%            & \textbf{98.7\%} & \underline{98.2\%} & 50.2\% & 90.5\%      & 73.8\%     & 95.7\%              & 97.7\% \\
\multicolumn{1}{l}{GSM8K}                         & \underline{77.9\%}    & 75.1\%            & \textbf{79.5\%} & 75.7\% & 28.2\% & 59.4\%      & 52.9\%     & 77.7\%             & 71.4\% \\
\multicolumn{1}{l}{AddSub}                           & 91.9\%    & 89.9\%            & \textbf{94.9\%} & 92.4\% & 55.4\% & 89.6\%      & 64.8\%     & \underline{94.9\%}             & 91.5\% \\
\multicolumn{1}{l}{SingleEq}                            & 95.7\%    & 97.0\%            & \textbf{98.4\%} & \underline{97.6\%} & 53.0\% & 92.3\%      & 67.5\%     & \underline{97.6\%}             & 96.3\% \\
\multicolumn{1}{l}{SVAMP}                            & \underline{82.7\%}    & 82.1\%            & 80.8\% & \textbf{84.8\%} & 44.9\% & 76.9\%      & 59.5\%     & 82.1\%             & 81.6\% \\
\rowcolor[HTML]{DEDEDE}
\multicolumn{1}{l}{Average} & 88.5\%    & 87.6\%            & \textbf{90.5\%} & \underline{89.7\%} & 46.3\% & 81.7\%      & 63.7\%     & 89.6\%             & 87.7\% \\ \thickhline
% \end{tabularx}
\end{tabular}}
\captionsetup{font=small, width=0.9\linewidth}
\caption{We compare the results across various arithmetic problem datasets using GPT-3.5. $^\dagger$ means whether the method uses external tools or not.}
\label{tab:gpt3.5_arithmetic_perf}
\end{table*}

%% file: tables/open_source_model.tex
\begin{table*}[h]
\centering
\small
\resizebox{16cm}{!}{
\begin{tabular}{llccccccc}
\thickhline
\textbf{Model name}          & \textbf{Dataset}                    & \textbf{Tool-Integrated} & \textbf{MU\_1}   & \textbf{MU\_2}   & \textbf{BeyondX\_3}   & \textbf{BeyondX\_4}   & \textbf{BeyondX\_5}   \\ \hline
Mistral-7B (Base)         & -                          & -               & 7.3\%  & 5.5\%  & 3.1\%  & 1.3\%  & 8.0\%  \\
Arithmo-Mistral-7B  & MetaMathQA \&MathInstruct & \xmark               & 43.5\% & 26.8\% & 12.9\% & 9.5\%  & 8.9\%  \\
MetaMath-Mistral-7B & MetaMathQA                 & \xmark               & 46.7\% & 51.0\% & 15.5\% & \underline{15.2\%} & \textbf{16.1\%} \\
WizardMath-Mistral-7B  & Not released               & \xmark               & \underline{65.2\%} & \underline{58.7\%} & \underline{20.6\%} & 13.3\% & \underline{14.3\%} \\
MMIQC-Mistral-7B    & MMIQC                      & \xmark               & 29.6\% & 12.4\% & 8.2\%  & 7.0\%  & 6.2\%  \\
MAmmoTH-Mistral-7B  & MathInstruct               & \cmark               & 46.8\% & 23.7\% & 4.6\%  & 4.4\%  & 4.5\%  \\
MAmmoTH2-Mistral-7B         & WebInstruct                & \cmark               & \textbf{69.0\%} & \textbf{68.0\%} & \textbf{25.8\%} & \textbf{16.5\%} & 9.8\%  \\
OpenMath-Mistral-7B & OpenMathInstruct-1         & \cmark               & 12.1\% & 5.3\%  & 5.7\%  & 1.9\%  & 0.9\%  \\ \thickhline
\end{tabular}}
\captionsetup{font=small, width=0.9\linewidth}
\caption{Experiment results of open-source math models that are fine-tune on Mistral-7B base model across various algebra problem datasets under Zero-shot-CoT setting.}
\label{tab:open_source_perf}
\end{table*}

%% file: tables/hyperparameters.tex
\begin{table*}[h]
\centering
\small
\resizebox{16cm}{!}{
\begin{tabular}{lll}
\thickhline
Model                 & Version                                          & Generation Setup                                      \\ \hline
\multicolumn{3}{c}{\textit{Close-source}}                                                                                                 \\ \hline
GPT-3.5               & gpt-3.5-turbo-1106                               & temperature = 0, max tokens = 2048                    \\
GPT-4                 & gpt-4-1106-preview                               & temperature = 0, max tokens = 2048                    \\
Gemini-Pro            & gemini-1.0-pro                                   & temperature = 0, max tokens = 2048                    \\ \hline
\multicolumn{3}{c}{\textit{Open-source}}                                                                                                  \\ \hline
Mistral-7B            & mistralai/Mistral-7B-v0.1                        & do\_sample=True, temperature=0.001, max tokens = 2048 \\
Arithmo-Mistral-7B    & akjindal53244/Arithmo-Mistral-7B                 & do\_sample=True, temperature=0.001, max tokens = 2048 \\
MetaMath-Mistral-7B   & meta-math/MetaMath-Mistral-7B                    & do\_sample=True, temperature=0.001, max tokens = 2048 \\
WizardMath-Mistral-7B & WizardLM/WizardMath-7B-V1.1                      & do\_sample=True, temperature=0.001, max tokens = 2048 \\
MMIQC-Mistral-7B      & Vivacem/Mistral-7B-MMIQC                         & do\_sample=True, temperature=0.001, max tokens = 2048 \\
MAmmoTH-Mistral-7B    & TIGER-Lab/MAmmoTH-7B-Mistral                     & do\_sample=True, temperature=0.001, max tokens = 2048 \\
MAmmoTH2-Mistral-7B   & TIGER-Lab/MAmmoTH2-7B                            & do\_sample=True, temperature=0.001, max tokens = 2048 \\
OpenMath-Mistral-7B   & nvidia/OpenMath-Mistral-7B-v0.1-hf               & do\_sample=True, temperature=0.001, max tokens = 2048 \\ \thickhline
\end{tabular}}
\captionsetup{font=small, width=0.9\linewidth}
\caption{Model names, versions, and generating setups for various LLMs.}
\label{tab:hyperparameter}
\end{table*}

%% file: tables/dataset_error_rate.tex
% \usepackage{multirow}
\begin{table}[h]
\centering
\resizebox{7.5cm}{!}{%
\begin{tabular}{lcccc}
\thickhline
\textbf{Source Dataset} & \textbf{Avg \#unknowns} & \textbf{Size} & \textbf{Avg \#words} & \textbf{Unreasonable Problem Rate} \\ \hline

ALG514  & 3 & 97 & 68.8 & 25.4\% (33/130) \\

 & 4 & 78 & 96.6 & 20.4\% (20/98) \\

 & 5 & 55 & 128.4 & 11.3\% (7/62) \\ \hline

DRAW-1K & 3 & 97 & 61.5 & 30.2\% (42/139) \\

 & 4 & 80 & 92.4 & 25.2\% (27/107) \\

 & 5 & 57 & 109.2 & 14.9\% (10/67) \\ \thickhline

\end{tabular}}
\captionsetup{font=small, width=0.9\linewidth}
\caption{Statistics of Proposed dataset. Unreasonable Problem Rate means \#Unreasonable Problem/ \#Total Human Seen Problem}

\label{tab:error_rate_dataset}

\end{table}

%% file: tables/different_shot.tex
\begin{table}[ht!]
\small
\begin{tabularx}{\linewidth}{XXXX}
\thickhline
\textbf{K-Shot}  & \textbf{3-shot} & \textbf{5-shot} & \textbf{8-shot} \\ \hline
\multicolumn{4}{c}{\textit{Single}}         \\ \hline
ALG514  & 92.3\% & \underline{92.3\%} & \textbf{95.6\%} \\
DRAW-1K  & 82.6\% & \underline{85.2\%} & \textbf{90.7\%} \\
AsDiv   & 79.7\% & \underline{81.7\%} & \textbf{85.5\%} \\
HMWP    & 34.5\% & \underline{37.7\%} & \textbf{38.0\%} \\
\rowcolor[HTML]{DEDEDE}Average & 72.3\% & \underline{74.2\%} & \textbf{77.5\%} \\ \hline
\multicolumn{4}{c}{\textit{Double}}         \\ \hline
ALG514  & 96.0\% & \underline{96.5\%} & \textbf{97.1\%} \\
DRAW-1K  & 84.9\% & \underline{85.6\%} & \textbf{85.6\%} \\
AsDiv   & \textbf{82.7\%} & 80.7\% & \underline{81.7\%} \\
HMWP    & 57.9\% & \underline{58.5\%} & \textbf{60.9\%} \\
\rowcolor[HTML]{DEDEDE}Average & \underline{80.4\%} & 80.3\% & \textbf{81.3\%} \\ \thickhline
\end{tabularx}
\captionsetup{font=small, width=0.9\linewidth}
\caption{Performance comparison of our method across different shots using GPT-3.5.}
\label{tab:different_shot}
\end{table}

%% file: tables/model_MU.tex
\begin{table*}[h]
\centering
\small
\begin{tabular}{l|l|cc|cccc|ccc}
% {p{1.0cm}p{1.0cm}p{1.0cm}p{1.0cm}p{1.6cm}}
% \begin{tabularx}{\linewidth}{XXXXXXXXXXX}
\thickhline
\multirow{2}{*}{Model} & \multirow{2}{*}{Setting} & \multicolumn{2}{c|}{Zero-shot} & \multicolumn{4}{c|}{Few-shot (Manual)} & \multicolumn{3}{c}{Few-shot (Automatic)} \\ \cline{3-11}
& & CoT & PS & CoT & PoT$^\dagger$ & EoT$^\dagger$ & DR$^\dagger$ & AG & AZ & Ours$^\dagger$ \\ \hline
\multirow{6}{*}{Gemini}
& \multicolumn{1}{l|}{MU\_1} & 51.0\% & 52.4\% & 67.9\% & 77.2\% & 46.9\% & 82.8\% & 62.1\% & \underline{84.8\%} & \textbf{86.2\%} \\
& \multicolumn{1}{l|}{MU\_2} & 67.1\% & 69.6\% & 51.3\% & 46.6\% & 36.6\% & \underline{92.3\%} & 32.5\% & 83.5\% & \textbf{93.1\%} \\
& \multicolumn{1}{l|}{BeyondX\_3} & 23.7\% & 23.7\% & 19.1\% & 18.6\% & 0.5\%  & \underline{86.1\%} & 14.9\% & 13.4\% & \textbf{95.9\%} \\
& \multicolumn{1}{l|}{BeyondX\_4} & 12.0\% & 12.0\% & 10.1\% & 18.4\% & 0.0\%  & \underline{80.4\%} & 12.7\% & 8.2\%  & \textbf{87.3\%} \\
& \multicolumn{1}{l|}{BeyondX\_5} & 9.8\%  & 12.5\% & 12.5\% & 10.7\% & 0.0\%  & \underline{65.2\%} & 8.9\%  & 11.6\% & \textbf{75.0\%} \\
\rowcolor[HTML]{DEDEDE}
& \multicolumn{1}{l|}{Average} & 32.7\% & 34.1\% & 32.2\% & 34.3\% & 16.8\% & \underline{81.3\%} & 26.2\% & 40.3\% & \textbf{87.5\%} \\ \hline
\multirow{6}{*}{GPT-4}
& \multicolumn{1}{l|}{MU\_1} & 88.3\% & 87.6\% & 86.2\% & 86.9\% & 36.9\% & 84.8\% & 87.6\% & \underline{89.0\%} & \textbf{95.2\%} \\
& \multicolumn{1}{l|}{MU\_2} & 92.3\% & \underline{92.4}\% & 86.3\% & 87.2\% & 32.3\% & 83.1\% & 76.3\% & 92.3\% & \textbf{96.3\%} \\
& \multicolumn{1}{l|}{BeyondX\_3} & 76.3\% & 72.7\% & 67.5\% & 73.7\% & 2.6\% & \underline{90.7\%} & 13.9\% & 86.1\% & \textbf{97.4\%} \\
& \multicolumn{1}{l|}{BeyondX\_4} & 29.7\% & 32.3\% & 31.0\% & 50.0\% & 0.6\% & \underline{83.5\%} & 1.9\%  & 41.8\% & \textbf{90.5\%} \\
& \multicolumn{1}{l|}{BeyondX\_5} & 21.4\% & 19.6\% & 17.0\% & 43.8\% & 1.8\% & \underline{60.7\%} & 0.9\%  & 28.6\% & \textbf{81.3\%} \\
\rowcolor[HTML]{DEDEDE} 
& \multicolumn{1}{l|}{Average} & 61.6\% & 60.9\% & 57.6\% & 68.3\% & 14.8\% & \underline{80.6\%} & 36.1\% & 67.5\% & \textbf{92.1\%} \\ \thickhline
\end{tabular}
\captionsetup{font=small, width=0.9\linewidth}
\caption{Experiment results across various unknowns using Gemini-Pro and GPT-4. $^\dagger$ means whether the method uses external tools or not.}
\label{tab:model_mu_perf}
\end{table*}

%% file: tables/model_main_table.tex
\begin{table*}[htb]
\centering
\small
% \begin{tabular}{cccccccccc}
% \begin{tabularx}{\linewidth}{XXXXXXXXXXX}
\begin{tabular}{l|l|cc|cccc|ccc}
\thickhline
\multirow{2}{*}{Model} & \multirow{2}{*}{Setting} & \multicolumn{2}{c|}{Zero-shot} & \multicolumn{4}{c|}{Few-shot (Manual)} & \multicolumn{3}{c}{Few-shot (Automatic)} \\ \cline{3-11}
& & CoT & PS & CoT & PoT$^\dagger$ & EoT$^\dagger$ & DR$^\dagger$ & AG & AZ & Ours$^\dagger$ \\ \hline
\multicolumn{11}{c}{\textit{Single}} \\ \hline
\multirow{5}{*}{Gemini}
& \multicolumn{1}{l|}{ALG514} & 55.0\% & 55.0\% & 82.4\% & 89.0\% & 52.8\% & 86.8\% & 70.3\% & \underline{89.0\%} & \textbf{94.5\%} \\
& \multicolumn{1}{l|}{DRAW-1K} & 44.4\% & 48.2\% & 43.3\% & 57.4\% & 37.0\% & \underline{75.9\%} & 48.2\% & \textbf{77.8\%} & 72.2\% \\
& \multicolumn{1}{l|}{AsDiv} & 47.9\% & 49.3\% & 57.8\% & \underline{75.0\%} & 60.6\% & 74.7\% & 49.3\% & 74.7\% & \textbf{77.5\%} \\
& \multicolumn{1}{l|}{HMWP} & 18.6\% & 17.0\% & 22.3\% & 15.1\% & 23.6\% & 23.1\% & 26.0\% & \underline{30.8\%} & \textbf{46.6\%} \\
\rowcolor[HTML]{DEDEDE} 
& \multicolumn{1}{l|}{Average} & 41.5\% & 42.3\% & 51.4\% & 59.1\% & 43.5\% & 65.1\% & 48.4\% & \underline{68.1\%} & \textbf{72.7\%} \\ \hline
\multirow{5}{*}{GPT-4}
& \multicolumn{1}{l|}{ALG514} & 92.3\% & 93.4\% & 92.3\% & 90.1\% & 39.9\% & 86.8\% & 93.4\% & \underline{93.4\%} & \textbf{98.9\%} \\
& \multicolumn{1}{l|}{DRAW-1K} & \underline{81.5\%} & 77.8\% & 75.9\% & 81.5\% & 31.9\% & 81.5\% & 77.8\% & 81.5\% & \textbf{88.9\%} \\
& \multicolumn{1}{l|}{AsDiv} & \underline{88.9\%} & 88.9\% & 88.9\% & 62.5\% & 11.3\% & 81.7\% & 84.1\% & 83.1\% & \textbf{90.1\%} \\
& \multicolumn{1}{l|}{HMWP} & 54.9\% & 52.8\% & 53.1\% & 22.0\% & 0.3\% & 31.3\% & 32.6\% & \textbf{55.4\%} & \underline{54.9\%} \\
\rowcolor[HTML]{DEDEDE}
& \multicolumn{1}{l|}{Average} & \underline{79.4\%} & 78.2\% & 77.5\% & 64.0\% & 20.8\% & 70.3\% & 72.0\% & 78.4\% & \textbf{83.2\%} \\ \hline
\multicolumn{11}{c}{\textit{Double}} \\ \hline
\multirow{5}{*}{Gemini}
& \multicolumn{1}{l|}{ALG514} & 69.3\% & 69.3\% & 53.9\% & 47.8\% & 39.7\% & \underline{94.8\%} & 33.3\% & 86.8\% & \textbf{95.3\%} \\
& \multicolumn{1}{l|}{DRAW-1K} & 61.0\% & 70.6\% & 43.8\% & 43.2\% & 27.4\% & \underline{84.9\%} & 30.1\% & 74.0\% & \textbf{87.0\%} \\
& \multicolumn{1}{l|}{AsDiv} & 77.1\% & 74.7\% & 63.9\% & 7.5\% & 7.2\% & \underline{85.5\%} & 49.4\% & 74.7\% & \textbf{84.3\%} \\
& \multicolumn{1}{l|}{HMWP} & 25.2\% & 32.2\% & 32.8\% & 26.9\% & 2.9\% & \underline{56.7\%} & 25.7\% & 43.9\% & \textbf{61.7\%} \\
\rowcolor[HTML]{DEDEDE} 
& \multicolumn{1}{l|}{Average} & 58.1\% & 61.7\% & 48.6\% & 31.3\% & 19.3\% & \underline{80.5\%} & 34.7\% & 69.8\% & \textbf{82.1\%} \\ \hline
\multirow{5}{*}{GPT-4}
& \multicolumn{1}{l|}{ALG514} & \underline{97.2\%} & 96.9\% & 91.5\% & 89.8\% & 32.4\% & 84.4\% & 80.1\% & 94.3\% & \textbf{98.8\%} \\
& \multicolumn{1}{l|}{DRAW-1K} & 78.1\% & 79.5\% & 71.2\% & 79.5\% & 32.1\% & 79.5\% & 65.1\% & \underline{86.3\%} & \textbf{89.0\%} \\
& \multicolumn{1}{l|}{AsDiv} & 86.8\% & 85.7\% & 83.5\% & 11.6\% & 2.4\% & 79.5\% & 39.6\% & \textbf{91.6\%} & \underline{90.4\%} \\
& \multicolumn{1}{l|}{HMWP} & 64.3\% & 64.9\% & 57.3\% & 56.3\% & 1.2\% & 57.9\% & 38.6\% & \underline{71.4\%} & \textbf{75.4\%} \\
\rowcolor[HTML]{DEDEDE}
& \multicolumn{1}{l|}{Average} & 81.6\% & 81.8\% & 75.9\% & 59.3\% & 17.0\% & 75.3\% & 55.8\% & \underline{85.9\%} & \textbf{88.4\%} \\ \thickhline
\end{tabular}
\captionsetup{font=small, width=0.9\linewidth}
\caption{Experiment results across various algebra problem datasets which include single and double unknown using Gemini-Pro and GPT-4. $^\dagger$ means whether the method uses external tools or not.}
\label{tab:model_main_perf}
\end{table*}

%% file: tables/MWP_generation_prompt.tex
\begin{table*}[h]
\small
\begin{tabularx}{\linewidth}{X}
\thickhline
\textbf{Instruction for Automatic Generation of Multiple Unknown Algebra Problems} \\ \hline
1. Write down each variable meaning in a system of equations by understanding the relation between a system of equations and a math word problem. \\
2. Introduce a new variable based on the existing variable meaning. Then, assign a value to the new variable based on the solution of the existing system of equations.\\
3. Add new terms and new equations to the existing system of equations to generate a solvable three-unknown system of equations.\\
4. Introduce new statements that are related to the existing math word problem or modify the existing statement to rewrite the problem into a three-unknown math word problem based on the new system of equations and the same original scenario.\\
5. Formulate and rephrase the statements and scenario into a coherent and reasonable math word problem. \\ \thickhline
\end{tabularx}
\captionsetup{font=small, width=0.9\linewidth}
\caption{Actual instruction of algebra problem generation.}
\label{tab:mwp_generation_instruction}
\end{table*}

%% file: tables/MWP_solver_prompt.tex
\begin{table*}[h]
\small
\begin{tabularx}{\linewidth}{X}
\thickhline
\textbf{Instruction for Automatic Solver of Algebra Problems} \\ \hline
1. Determine what the question is asking. \\
2. Write down the relevant information in simple statements.\\
3. Assign symbols (must be an alphabetic character e.g., x, y, z etc.) to unknown values that must be found. \\
4. Determine how the statements relate to each other mathematically.\\
5. Give the equations only here, with each equation on a new line. \\ \thickhline
\end{tabularx}
\captionsetup{font=small, width=0.9\linewidth}
\caption{Actual instruction of algebra problem solver.}
\label{tab:mwp_solver_instruction}
\end{table*}

%% file: tables/direct_prompt_for_mwp_generation.tex
\begin{table*}[h]
\centering
\small
\begin{tabularx}{\linewidth}{X}

\thickhline

\rowcolor[HTML]{DEDEDE}
\textbf{Prompt:} \\
Generate a three-unknown math word problem and its corresponding system of equations.                                                                                             \\ \hline

\textbf{Example 1:}\\ Word Problem:\\ Three friends, Alice, Bob, and Carol, decide to pool their money to buy a gift for a friend. Alice contributes twice as much as Bob, while Carol contributes \$10 more than Bob. The total amount they collect is \$110. How much did each person contribute?\\ \\ System of Equations:\\ Let A represent the amount Alice contributes,B represent the amount Bob contributes, and C represent the amount Carol contributes. Based on the information given:\\ 1. A = 2B (Alice contributes twice as much as Bob)\\ 2. C = B + 10 (Carol contributes \$10 more than Bob)\\ 3. A + B + C = 110 (The total amount collected is \$110) \\ \hline

\textbf{Example 2:}\\ Word Problem:\\ During a school fundraiser, three students, Emily, Daniel, and Fiona, sold raffle tickets. Emily sold 5 more tickets than Daniel. Fiona sold twice as many tickets as Emily. Together, they sold a total of 155 tickets. How many tickets did each student sell?\\ \\ System of Equations:\\ Let E represent the number of tickets Emily sold, D represent the number of tickets Daniel sold, and F represent the number of tickets Fiona sold. Based on the information given:\\ 1. E = D + 5 (Emily sold 5 more tickets than Daniel)\\ 2. F = 2E (Fiona sold twice as many tickets as Emily)\\ 3. E + D + F = 155 (Together, they sold a total of 155 tickets) \\ \hline

\textbf{Example 3:}\\ Word Problem:\\ In a small bakery, three types of pastries are made daily: croissants, muffins, and scones. The number of croissants made is twice the number of muffins. The bakery made 30 more scones than muffins. At the end of the day, they counted a total of 210 pastries. How many of each type of pastry were made?\\ \\ System of Equations:\\ Let C represent the number of croissants, M represent the number of muffins, and S represent the number of scones. Based on the information given:\\ 1. C = 2M (The number of croissants is twice the number of muffins)\\ 2. S = M + 30 (There are 30 more scones than muffins)\\ 3. C + M + S = 210 (The total number of pastries is 210) \\ \thickhline

\end{tabularx}
\captionsetup{font=small, width=0.9\linewidth}
\caption{Examples of multiple unknown algebra problem and corresponding system of equations using simple prompt on GPT-4.}

\label{tab:direct_mwp_generation}

\end{table*}

%% file: tables/equation_aware_mwp_generation.tex
\begin{table*}[h]
\centering
\small
\resizebox{14.5cm}{!}{%
\begin{tabularx}{\linewidth}{X}
\thickhline
\rowcolor[HTML]{DEDEDE}
\textbf{Prompt 1:} \\
Let's generate a three-unknown math word problem and its according system of equations. First, expand the two-unknown system of equations to three-unknown. Then, transfer the system of equations into according math word problem.\\
System of Equations:\\
0.01 * 35.0 * x + 0.01 * 15.0 * y = 0.01 * 27.0 * ( 200.0 )\\
y = ( 200.0 - x ) \\ \hline
\textbf{Response 1:} \\
Expanded System of Equations:\\
0.01 * 35.0 * x + 0.01 * 15.0 * y + 0.01 * 25.0 * z = 0.01 * 27.0 * 200.0\\
y + z = 200.0 - x\\
x - y = z - 50.0\\
Math Word Problem:\\
A company is investing in three different projects. The first project is expected to yield a 35\% return, the second a 15\% return, and the third a 25\% return. The total investment is \$200 and the total expected return is 27\% of the total investment. If the company invests \$50 more in the first project than in the third, and the total investment in the second and third projects equals the remaining amount after investing in the first project, how much is invested in each project?\\
\textbf{\color{red}{Source Problem:}}\\
An alloy containing 15 \% brass is to be combined with an alloy containing 35 \% brass to form an alloy containing 27 \% brass. How much of each alloy should be combined to make 200 pounds of the 27 \% brass alloy? \\ \hline
\rowcolor[HTML]{DEDEDE}
\textbf{Prompt 2:} \\
Let's generate a three-unknown math word problem and its according system of equations. First, expand the two-unknown system of equations to three-unknown. Then, transfer the system of equations into according math word problem.\\
System of Equations:\\
6.0 * 0.01 * x + 9.0 * 0.01 * y = 684.0\\
x + y = 10000.0 \\ \hline
\textbf{Response 2:} \\
Expanded System of Equations:\\
6.0 * 0.01 * x + 9.0 * 0.01 * y + 5.0 * 0.01 * z = 684.0\\
x + y + z = 10000.0\\
x - y = 2000.0\\
Math Word Problem:\\
A company is investing in three different projects. The first project is expected to yield a 6\% return, the second a 9\% return, and the third a 5\% return. The total investment is \$10,000 and the total expected return is \$684. If the company invests \$2000 more in the first project than in the second, how much is invested in each project?\\
\textbf{\color{red}{Source Problem:}}\\
Suppose you invested 10,000 dollars, part at 6 \% annual interest and the rest at 9 \% annual interest. If you received 684 dollars in interest after one year, how much did you invest at 6 \% annual interest and at 9 \% annual interest? \\ \hline
\rowcolor[HTML]{DEDEDE}
\textbf{Prompt 3:} \\
Let's generate a three-unknown math word problem and its according system of equations. First, expand the two-unknown system of equations to three-unknown. Then, transfer the system of equations into according math word problem.\\
System of Equations:\\
7.0 * x + 5.0 * y = 146.0\\
x + y = 24.0 \\ \hline
\textbf{Response 3:} \\
Expanded System of Equations:\\
7.0 * x + 5.0 * y + 3.0 * z = 146.0\\
x + y + z = 24.0\\
x - y = 2.0\\
Math Word Problem:\\
A store sells three types of fruits: apples, bananas, and cherries. Apples cost \$7 per pound, bananas cost \$5 per pound, and cherries cost \$3 per pound. The store sold a total of 24 pounds of these fruits and made \$146. If the store sold 2 more pounds of apples than bananas, how many pounds of each type of fruit were sold?\\
\textbf{\color{red}{Source Problem:}}\\
Samuel has found that his new car gets 7 miles per gallon on the highway and 5 miles per gallon in the city. He recently drove 146 miles on 24 gallons of gasoline. How many miles did he drive on the highway? How many miles did he drive in the city? \\ \hline
\rowcolor[HTML]{DEDEDE}
\textbf{Prompt 4:} \\
Let's generate a three-unknown math word problem and its according system of equations. First, expand the two-unknown system of equations to three-unknown. Then, transfer the system of equations into according math word problem.\\
System of Equations:\\
5.0 + y = x\\
4.0 * x + 3.0 * y = 97.0 \\ \hline
\textbf{Response 4:} \\
Expanded System of Equations:\\
5.0 + y + z = x\\
4.0 * x + 3.0 * y - z = 97.0\\
x - y = z + 2.0\\
Math Word Problem:\\
Three friends, Alice, Bob, and Charlie, are playing a game. The score of Alice is 5 more than Bob's score. When Alice's score is multiplied by 4 and added to 3 times Bob's score, it equals 97 minus Charlie's score. If Alice's score minus Bob's score equals Charlie's score plus 2, what are the individual scores of Alice, Bob, and Charlie?\\
\textbf{\color{red}{Source Problem:}}\\
One number is 5 more than another number. Four times the larger number plus 3 times the smaller is 97. Find the numbers. \\ \thickhline
\end{tabularx}}
\captionsetup{font=small, width=0.9\linewidth}
\caption{Examples of multiple unknown algebra problem generation, corresponding system of equations and source problem using equation-aware prompt on GPT-4.}
\label{tab:equation_aware_mwp_generation_part1}
\end{table*}

%% file: tables/error_cases_mwp_generation.tex
\begin{table*}[htb]
\centering
\small
\begin{tabularx}{\linewidth}{X}
\thickhline
\textbf{Case 1:} How much 1 \% boric acid solution, 5 \% boric acid solution, and 2\% boric acid solution are needed to make 40 mL of a solution that is 1.1\% boric acid? The sum of the amounts of the 5\% solution and the 2\% solution equals the difference between 40 mL and the amount of the 1\% solution. Also, the sum of the amounts of the 1\% solution and the 5\% solution equals twice the amount of the 2\% solution plus 10 mL.             \\
\begin{tabular}[c]{@{}l@{}}\textbf{System of Equations:}\\ ( 0.01 * 1.0 ) * x + ( 0.01 * 5.0 ) * y + ( 0.01 * 2.0 ) * z  = 1.10 (40 * 0.0275)\\ y + z  = 40.0 - x\\ x + y  = 2.0 * z + 10.0\end{tabular}                                                                                                                                                                                                                                                   \\
\textbf{Correct Question:} How much 1 \% boric acid solution, 5 \% boric acid solution, and 2\% boric acid solution are needed to make 40 mL of a solution that is \textbf{\color{red}2.75\%} boric acid? The sum of the amounts of the 5\% solution and the 2\% solution equals the difference between 40 mL and the amount of the 1\% solution. Also, the sum of the amounts of the 1\% solution and the 5\% solution equals twice the amount of the 2\% solution plus 10 mL. \\ \hline
\textbf{Case 2:} If the price of copper is 65 cents per pound, the price of zinc is 30 cents per pound, and the price of nickel is 50 cents per pound, how many pounds of copper, zinc, and nickel should be mixed to make 80 pounds of an alloy selling for 45.625 cents per pound? The sum of the pounds of zinc and nickel equals 80 minus the pounds of copper.                                                                                         \\
\begin{tabular}[c]{@{}l@{}}\textbf{System of Equations:}\\ 65.0 * x + 30.0 * y + 50.0 * z  = 3650.0\\ y + z  = 80.0 - x\\ x + y + z  = 80.0\end{tabular}                                                                                                                                                                                                                                                                                                    \\
\textbf{Correct Question:} If the price of copper is 65 cents per pound, the price of zinc is 30 cents per pound, and the price of nickel is 50 cents per pound, how many pounds of copper, zinc, and nickel should be mixed to make 80 pounds of an alloy selling for 45.625 cents per pound? \textbf{\color{red}The sum of the pounds of zinc and nickel equals 7 times the pounds of copper. (create new math relation)}                                                     \\ \hline
\textbf{Case 3:} Three cars are travelling in opposite directions. After two hours, they are 372 km apart. The first car is the slowest, the second car travels 6 kph faster than the first car, and the third car travels 4 kph slower than the first car. Find the average speed of each car.                                                                                                                                                             \\
\begin{tabular}[c]{@{}l@{}}\textbf{System of Equations:}\\ 2.0 * ( x + y + z )  = 372.0\\ y  = x + 6.0\\ z  = x - 81.0\end{tabular}                                                                                                                                                                                                                                                                                                                         \\
\textbf{Correct Question: \color{red}Cannot Expand}                                                                                                                                                                                                                                                                                                                                                                                                                    \\ \hline
\textbf{Case 4:} Sam, Masc, and Alex are friends. Masc is 7 years older than Sam and Alex is 5 years older than Masc. The sum of their ages is 49. What are their ages?                                                                                                                                                                                                                                                                                     \\
\begin{tabular}[c]{@{}l@{}}\textbf{System of Equations:}\\ y + x + z  = 37.0\\ y  = x + 7.0\\ z  = y - 7.0\end{tabular}                                                                                                                                                                                                                                                                                                                                     \\
\textbf{Correct Question:} Sam, Masc, and Alex are friends. Masc is 7 years older than Sam and Alex is \textbf{\color{red}7 years younger} than Masc. The sum of their ages is \textbf{\color{red}37}. What are their ages?                                                                                                                                                                                                                                                                         \\ \thickhline
\end{tabularx}
\captionsetup{font=small, width=0.9\linewidth}
\caption{Error cases of multiple unknown algebra problem generation, corresponding system of equations, and correct question.}
\label{tab:error_cases_mwp_generation}
\end{table*}

%% file: tables/error_cases_mwp_solver.tex
\begin{table*}[htb]
\centering
\small
\resizebox{14.5cm}{!}{%
\begin{tabularx}{\linewidth}{X}
\thickhline
\rowcolor[HTML]{DEDEDE}
\textbf{E1: Fewer or More Equation in the Set} \\
\textbf{Input:} \\
\{Instruction\}             \\
\{Demos\} \\
Question: The Rocket Coaster has expanded and now has 20 cars: some hold 4 people, some hold 6 people, and some hold 2 people. Altogether, there is room for 82 people. It is also known that the number of 4-passenger cars plus the number of 6-passenger cars equals three times the number of 2-passenger cars. How many of each type of car are there? \\ \hline
\textbf{Output:} \\
The system of equations for the given word problem is as follows:\\
1- The question is asking for the number of each type of car.\\
2- Relevant information:\\
\quad - Total number of cars: 20\\
\quad - Total capacity: 82 people\\
\quad - Relationship between the number of cars: 4-passenger + 6-passenger = 3 * 2-passenger\\
3- Assign symbols:\\
\quad Let a = number of 4-passenger cars\\
\quad Let b = number of 6-passenger cars\\
\quad Let c = number of 2-passenger cars\\
4- Mathematical relationships:\\
\quad - 4a + 6b + 2c = 82 (total capacity)\\
\quad - a + b = 3c (relationship between the number of cars)\\
5- System of equations:\\
\quad 4a + 6b + 2c = 82\\
\quad a + b = 3c \\ \hline
\rowcolor[HTML]{DEDEDE}
\textbf{E2: Wrong equation (meaning the coefficients or constant terms are wrong)} \\
\textbf{Input:} \\
\{Instruction\}             \\
\{Demos\} \\
Question: In a family, there are 3 cars. In a particular week, the sum of the average miles per gallon obtained by the three cars is 75. The first car has consumed 40 gallons, the second car has consumed 20 gallons, and the third car has consumed 10 gallons. This resulted in a total of 1700 miles driven by the three cars combined. Additionally, the sum of the average gas mileage of the first car and the third car is 15 more than twice the average gas mileage of the second car. What was the average gas mileage obtained by each of the 3 cars that week?\\ \hline
\textbf{Output:} \\
1- The question is asking for the average gas mileage obtained by each of the 3 cars in a particular week.\\
2- Relevant information:\\
\quad - Total sum of the average miles per gallon obtained by the three cars: 75\\
\quad - Gallons consumed by the first car: 40\\
\quad - Gallons consumed by the second car: 20\\
\quad - Gallons consumed by the third car: 10\\
\quad - Total miles driven by the three cars: 1700\\
\quad - Relationship between the average gas mileage of the cars\\
3- Assign symbols:\\
\quad Let a = average gas mileage of the first car\\
\quad Let b = average gas mileage of the second car\\
\quad Let c = average gas mileage of the third car\\
4- Mathematical relationships:\\
\quad - (40/a) + (20/b) + (10/c) = 75\\
\quad - 40a + 20b + 10c = 1700\\
\quad - (a + c) = 15 + 2b\\
5- System of equations:\\
\quad (40/a) + (20/b) + (10/c) = 75\\
\quad 40a + 20b + 10c = 1700\\
\quad (a + c) = 15 + 2b\\
\hline
\rowcolor[HTML]{DEDEDE}
\textbf{E3: Wrong Extraction Form of System of Equations} \\
\textbf{Input:} \\
\{Instruction\}             \\
\{Demos\} \\   
Question: The Drama Club sold 1,700 tickets for the end-of-year-performance. Admission prices were 12 dollars for adults, 6 dollars for students, and 10 dollars for seniors. The total amount collected at the box office was 18,200 dollars. It is known that twice the number of student tickets plus the number of adult tickets equals three times the number of senior tickets plus 1200. How many students, adults, and seniors attended the play?\\ \hline
\textbf{Output:} \\
The system of equations for the given word problem is as follows:\\
1. Let a be the number of adult tickets sold, s be the number of student tickets sold, and r be the number of senior tickets sold.\\
2. Mathematical relationship:\\
\quad - a + s + r = 1700 (total tickets sold)\\
\quad - 12a + 6s + 10r = 18200 (total revenue collected)\\
\quad - 2s + a = 3r + 1200 (number of tickets relationship)\\
The system of equations is:\\
\quad a + s + r = 1700\\
\quad 12a + 6s + 10r = 18200\\
\quad 2s + a = 3r + 1200\\ \thickhline
\end{tabularx}}
\captionsetup{font=small, width=0.9\linewidth}
\caption{Different type of error example of Formulate-and-Solve on GPT-3.5.}
\label{tab:error_example_1}
\end{table*}

%% file: tables/full_example.tex
\begin{table*}[htb]
\centering
\small
\begin{tabularx}{\linewidth}{X}
\thickhline
\rowcolor[HTML]{DEDEDE}
\textbf{Three-Unknown} \\
\textbf{Input:} \\
\{Instruction\}             \\
\{Demos\} \\
Question: A chemist has three solutions: one that is 18 \% alcohol, one that is 50 \% alcohol, and one that is 10 \% alcohol. He wants to make 100 liters of a solution that is 26 \% alcohol. Also, the amount of the 18 \% alcohol solution plus the amount of the 50 \% alcohol solution he uses equals four times the amount of the 10 \% alcohol solution. How many liters of each solution should he add? \\ \hline
\textbf{Output:} \\
The system of equations for the given word problem is as follows: \\
1- The question is asking for the number of liters of each solution the chemist should add. \\
2- Relevant information: \\
\quad - Total liters of solution: 100 \\
\quad - Desired alcohol concentration: 26\% \\
\quad - Relationship between the amount of solutions: amount of 18\% solution + amount of 50\% solution = 4 * amount of 10\% solution \\
3- Assign symbols: \\
\quad Let a = liters of 18\% alcohol solution \\
\quad Let b = liters of 50\% alcohol solution \\
\quad Let c = liters of 10\% alcohol solution \\
4- Mathematical relationships: \\
\quad - a + b + c = 100 (total liters of solution) \\
\quad - 0.18a + 0.50b + 0.10c = 0.26 * 100 (desired alcohol concentration) \\
\quad- a + b = 4c (relationship between the amount of solutions) \\
5- System of equations: \\
\quad a + b + c = 100 \\
\quad 0.18a + 0.50b + 0.10c = 26 \\
\quad a + b = 4c \\ 
\hline
\rowcolor[HTML]{DEDEDE}
\textbf{Four-Unknown} \\
\textbf{Input:} \\
\{Instruction\}             \\
\{Demos\} \\
Question: In a laboratory, a chemist needs to prepare 300 milliliters of a 20 \% HCl solution. She has available a 10 \% HCl solution, a 30 \% HCl solution, a 20 \% HCl solution, and a 40 \% HCl solution. She knows that the volume of the 30 \% HCl solution plus twice the volume of the 10 \% HCl solution equals five times the volume of the 20 \% HCl solution plus 100 milliliters. The volume of the 30 \% HCl solution plus the volume of the 10 \% HCl solution plus the volume of the 20 \% HCl solution equals four times the volume of the 40 \% HCl solution plus 50 milliliters. How many milliliters of each solution should she use?                                                                                        \\ \hline

\textbf{Output:} \\
The system of equations for the given word problem is as follows: \\
1- The question is asking for the volume of each solution to be used. \\
2- Relevant information: \\
\quad - Total volume needed: 300 milliliters \\
\quad - Mathematical relationships between the volumes of the solutions \\
3- Assign symbols:\\
\quad Let x be the volume of the 10\% HCl solution. \\
\quad Let y be the volume of the 30\% HCl solution. \\
\quad Let z be the volume of the 20\% HCl solution. \\ 
\quad Let w be the volume of the 40\% HCl solution. \\
4- Mathematical relationships: \\
\quad x + y + z + w = 300 (total volume needed)\\
\quad 0.1x + 0.3y + 0.2z + 0.4w = 0.2 * 300 (20\% HCl solution)\\
\quad y + 2x = 5z + 100 (relationship 1)\\
\quad y + x + z = 4w + 50 (relationship 2)\\
5- System of equations:\\
\quad x + y + z + w = 300\\
\quad 0.1x + 0.3y + 0.2z + 0.4w = 60\\
\quad y + 2x - 5z = 100\\
\quad y + x + z - 4w = 50
\\ \thickhline
\end{tabularx}
\captionsetup{font=small, width=0.9\linewidth}
\caption{Full output example of Formulate-and-Solve on GPT-3.5 from three and four-unknown problem.}
\label{tab:full_example_1}
\end{table*}

\begin{table*}[htb]
\centering
\small
\begin{tabularx}{\linewidth}{X}
\thickhline
\rowcolor[HTML]{DEDEDE}
\textbf{Five-Unknown} \\
\textbf{Input:} \\
\{Instruction\}             \\
\{Demos\} \\   
Question: A chemist has five solutions: one that is 39 \% alcohol, one that is 50 \% alcohol, one that is 12 \% alcohol, one that is 5 \% alcohol, and one that is 3 \% alcohol. He wants to make 140 liters of a solution that is 27.5 \% alcohol. Also, the amount of the 39 \% alcohol solution plus the amount of the 50 \% alcohol solution he uses equals four times the amount of the 12 \% alcohol solution. The amount of the 39 \% alcohol solution plus the amount of the 50 \% alcohol solution plus the amount of the 12 \% alcohol solution equals five times the amount of the 5 \% alcohol solution. The amount of the 39 \% alcohol solution plus the amount of the 50 \% alcohol solution plus the amount of the 12 \% alcohol solution plus the amount of the 5 \% alcohol solution equals six times the amount of the 3 \% alcohol solution. How many liters of each solution should he add?\\ \hline
\textbf{Output:} \\
I can help you with that. Let's start by writing down the relevant information in simple statements:\\
1- The question is asking for the amount of each solution the chemist should add to make 140 liters of a solution that is 27.5\% alcohol. \\
2- Relevant information: \\
\quad - The chemist has five solutions: 39\% alcohol, 50\% alcohol, 12\% alcohol, 5\% alcohol, and 3\% alcohol. \\
\quad - He wants to make 140 liters of a solution that is 27.5\% alcohol.\\
\quad - The amount of the 39\% alcohol solution plus the amount of the 50\% alcohol solution he uses equals four times the amount of the 12\% alcohol solution.\\
\quad - The amount of the 39\% alcohol solution plus the amount of the 50\% alcohol solution plus the amount of the 12\% alcohol solution equals five times the amount of the 5\% alcohol solution.\\
\quad - The amount of the 39\% alcohol solution plus the amount of the 50\% alcohol solution plus the amount of the 12\% alcohol solution plus the amount of the 5\% alcohol solution equals six times the amount of the 3\% alcohol solution.\\
3- Assign symbols:\\
\quad Let a = amount of 39\% alcohol solution\\
\quad Let b = amount of 50\% alcohol solution\\
\quad Let c = amount of 12\% alcohol solution\\
\quad Let d = amount of 5\% alcohol solution\\
\quad Let e = amount of 3\% alcohol solution\\
4- Mathematical relationships:\\
\quad a + b + c + d + e = 140\\
\quad 0.39a + 0.50b + 0.12c + 0.05d + 0.03e = 0.275 * 140\\
\quad a + b = 4c\\
\quad a + b + c = 5d\\
\quad a + b + c + d = 6e\\
5- System of equations:\\
\quad a + b + c + d + e = 140\\
\quad 0.39a + 0.50b + 0.12c + 0.05d + 0.03e = 38.5\\
\quad a + b - 4c = 0\\
\quad a + b + c - 5d = 0\\
\quad a + b + c + d - 6e = 0 \\ \thickhline
\end{tabularx}
\captionsetup{font=small, width=0.9\linewidth}
\caption{Full output example of Formulate-and-Solve on GPT-3.5 from five-unknown problem.}
\label{tab:full_example_3}
\end{table*}